\documentclass{article}

    \PassOptionsToPackage{numbers, compress}{natbib}
\usepackage[preprint]{neurips_2025}

\usepackage[utf8]{inputenc} % allow utf-8 input
\usepackage[T1]{fontenc}    % use 8-bit T1 fonts
\usepackage{hyperref}       % hyperlinks
\usepackage{url}            % simple URL typesetting
\usepackage{booktabs}       % professional-quality tables
\usepackage{amsfonts}       % blackboard math symbols
\usepackage{nicefrac}       % compact symbols for 1/2, etc.
\usepackage{microtype}      % microtypography
\usepackage{xcolor}         % colors

\usepackage{float}
\usepackage{microtype}
\usepackage{graphicx}
\usepackage{booktabs}
\usepackage{hyperref}
\usepackage{subcaption}
\usepackage{pifont}

\usepackage{amsmath}
\usepackage{amssymb}
\usepackage{mathtools}
\usepackage{amsthm}
\usepackage{spverbatim}
\usepackage[framemethod=TikZ]{mdframed}
\usepackage{graphicx}
\usepackage{booktabs}
\usepackage{wrapfig}
\usepackage{tabularx}
\usepackage{longtable}
\usepackage{titletoc}
\usepackage[capitalize,noabbrev]{cleveref}
\theoremstyle{plain}

\theoremstyle{definition}

\theoremstyle{remark}

\usepackage[textsize=tiny]{todonotes}
\usepackage[noamssymbols,upint]{newtxmath} 
\hypersetup{
    colorlinks=true,
    linkcolor=red,
    citecolor=blue,
    filecolor=magenta,      
    urlcolor=blue,
    linktocpage}
\usepackage[toc,page,header]{appendix}
\usepackage{minitoc}
\makeatletter
\newenvironment{smalleralign}[1][\small]
 {\par\nopagebreak\leavevmode\vspace*{-\baselineskip}%
  \skip0=\abovedisplayskip
  #1%
  \def\maketag@@@##1{\hbox{\m@th\normalfont\normalsize##1}}%
  \abovedisplayskip=\skip0
  \align}
 {\endalign\ignorespacesafterend}
\makeatother
\usepackage{algorithm}
\usepackage{algorithmic}

\usepackage{multirow}
\usepackage{booktabs}

\title{Natural Language Reinforcement Learning}

\author{%
\textbf{Xidong Feng$^{1}$\footnotemark[1]~~\footnotemark[2]~, Bo Liu$^{2}$\footnotemark[1]~~\footnotemark[2]~, Yan Song$^{1}$\footnotemark[2]~, Haotian Fu$^{3}$\footnotemark[2]~, Ziyu Wan$^{4}$\footnotemark[2]~, } \\
\textbf{Girish A. Koushik$^{5}$}, \textbf{Zhiyuan Hu$^2$}, \textbf{Mengyue Yang$^{6}$,  } 
 \textbf{Ying Wen$^4$}, \textbf{Jun Wang$^1$} \\ 
$^1$University College London, $^2$National University of Singapore, \\
$^3$Brown University, $^4$Shanghai Jiao Tong University, $^5$University of Surrey, $^6$University of Bristol \\
\texttt{\{fengxidongwh\}@gmail.com}
}

\begin{document}
\renewcommand{\thefootnote}{\fnsymbol{footnote}}
\footnotetext[1]{Equal contribution, correspondence to fengxidongwh@gmail.com}
\footnotetext[2]{Core contributors.}
\renewcommand*{\thefootnote}{\arabic{footnote}}
\renewcommand{\thefootnote}{\fnsymbol{footnote}}

\maketitle

\begin{abstract}
Artificial intelligence progresses towards the "Era of Experience," where agents are expected to learn from continuous, grounded interaction. We argue that traditional Reinforcement Learning (RL), which typically represents value as a scalar, can restrict agent's deep understanding of environments and hinders the active, deliberative learning crucial for navigating this new paradigm. To address the issue, we introduce Natural Language Reinforcement Learning (NLRL), a framework that extends RL principles into natural language counterparts. Central to NLRL is the Language Value Function (LVF), which redefines value as an interpretable linguistic narrative articulating the rationale behind an evaluation. NLRL further extends this concept to core RL components, including policy, the Bellman equation, and policy iteration. Leveraging recent advancements in Large Language Models (LLMs), NLRL can be practically implemented to achieve RL-like policy and value training through unsupervised environment interactions. Experiments over 4 multi-step agentic tasks demonstrate NLRL's effectiveness, efficiency, and its potential to foster deeper understanding and more active learning strategies.
\end{abstract}

\section{Introduction}

The pursuit of artificial intelligence is rapidly advancing towards the "Era of Experience," where agents are expected to learn and master complex tasks through continuous, grounded interaction with their environments, thereby moving beyond the inherent limitations of pre-existing human data \citep{silver2025welcome}. Central to this paradigm shift is the ability of agents to not only act but also to deeply understand and evaluate their experiences to inform future decisions. Traditional Reinforcement Learning (RL) \citep{sutton2018reinforcement} has laid a robust mathematical foundation for such decision-making, primarily through the concept of a value function, which estimates the cumulative reward of states or state-action pairs.

The conventional representation of value is scalar quantity \citep{sutton1999policy}, acting as highly compressed summaries of future trajectories. While computationally convenient for optimization, it fundamentally limits an agent's capacity for a deep, structural understanding of its experiences. This limitation becomes particularly salient when considering the vast, nuanced information embedded within an agent's interaction stream – information that a single numerical value struggles to encapsulate. For instance, why is a particular state valuable? Is it due to an immediate tactical advantage, a long-term strategic opportunity, a successful mitigation of a threat, or a combination thereof? A scalar value provides the "what" (how good) but not the "why." Without this crucial "why," the agent operates with a superficial grasp of its environment and the true drivers of long-term success. 

This superficial understanding, in turn, shapes the very nature of the learning process. While policy optimization methods like policy gradients \citep{williams1992simple, sutton1999policy} effectively reinforce successful behaviors, their efficacy is largely predicated on a "passive" discovery through sampling: the agent can only refine its strategy by first observing the outcomes of sampled actions, essentially waiting to stumble upon and then amplify beneficial action sequences. Without a richer, more interpretable grasp of why certain actions yield better outcomes, it is challenging to shift from passive sampling. It cannot engage in "active" or "deliberative" reasoning over the root causes of success or failure to proactively and systematically adjust its strategies. To truly unlock the potential of experiential learning and build agents capable of nuanced understanding, transparent reasoning, and more active, deliberative learning, we argue for a richer, more expressive representation of value function.

In this paper, we propose a novel approach, \textbf{Natural Language Reinforcement Learning} (NLRL), that directly addresses this challenge by re-envisioning the core components of RL within the domain of natural language. At the heart of NLRL is the concept of the Language Value Function (LVF). Instead of a simple scalar, the LVF redefines value as a rich, interpretable linguistic narrative. LVF allows an agent to articulate why a state or action is valuable, drawing justifications from textualized experiences and observations. This richer representation of value is crucial for learning actively from the diverse and often nuanced feedback characteristic of the "Era of Experience."

Building upon the Language Value Function, our NLRL framework further extends other key RL elements – including policy, Bellman equation, and policy iteration – into their language counterparts. We demonstrate how these components can be practically implemented with the recent success of large language models (LLMs). Our experiments in 4 multi-step agentic environments showcase the effectiveness, efficiency, and particularly potentials for more active learning within the NLRL framework compared to traditional RL approaches, paving the way for agents that learn not just to act, but to understand, reason, and actively improve. The codebase is \href{https://anonymous.4open.science/r/Natural-language-RL-1923}{available}.

\section{Preliminary of Reinforcement Learning}
\label{sec:preliminary}
RL models the decision-making problem as a Markov Decision Process (MDP), defined by the state space $\mathcal{S}$, action space $\mathcal{A}$, probabilistic transition function $P: \mathcal{S} \times \mathcal{A} \times \mathcal{S} \rightarrow[0,1]$, discount factor $\gamma \in[0,1)$ and reward function $r: \mathcal{S} \times \mathcal{A} \rightarrow\left[-R_{\max }, R_{\max }\right]$. The goal of RL aims to learn a policy $\pi: \mathcal{S} \times \mathcal{A} \rightarrow[0,1]$, which measures the action $a$'s probability given the state $s$: $\pi(a|s)=\operatorname{Pr}\left(A_t=a \mid S_t=s\right)$. In decision-making tasks, the optimal policy tends to maximize the expected discounted cumulative reward: $\pi^{*}(a|s)=\arg\max_{\pi}\mathbb{E}_{\pi}\big[\sum_{t=0}^{\infty} \gamma^t r\left(\mathrm{s}_t, \mathrm{a}_t\right)\big]$. The state-action and state value functions are two key concepts that evaluate states or state-action pairs by measuring the cumulative reward starting from them:
$Q_\pi(s_t,a_t)=\mathbb{E}_{(s,a)_{t+1: \infty} \sim P_\pi}\left[ \sum_{i=t}^\infty \gamma^{i-t} r\left(s_i,a_i\right) \mid s_t, a_t\right]$, $V_\pi(s_{t}) = \mathbb{E}_{a_{t}, (s,a)_{t+1: \infty} \sim P_\pi}\left[\sum_{i=t}^\infty \gamma^{i-t} r\left(s_i,a_i\right) \mid s_t\right]$, where $P_{\pi}$ is the trajectory distribution given the policy $\pi$ and dynamic transition $P$. Given the definition of $V_\pi(s_{t})$, the relationship between temporally adjacent state's value (e.g., $V(s_{t})$ and $V(s_{t+1})$) forms the Bellman expectation equation \citep{bellman1965dynamic}. For all states $s_{t} \in \mathcal{S}$, the one-step Bellman expectation equation is:
\begin{smalleralign}
\label{equ:one-step-bellman}
V_\pi(s_t)=\mathbb{E}_{a_t\sim\pi_\theta}[r(s_t,a_t)+\gamma \mathbb{E}_{s_{t+1}\sim p(s_{t+1}|s_t,a_t)}[V_\pi(s_{t+1})]], \forall s_{t} \in \mathcal{S},
\end{smalleralign}
% % \begin{smalleralign}
% % \label{equ:one-step-bellman}
% % V_\pi(s_t)=\mathbb{E}_{a_t\sim\pi_\theta}\Big[r(s_t,a_t)+\gamma \mathbb{E}_{s_{t+1}\sim p(s_{t+1}|s_t,a_t)}[V_\pi(s_{t+1})]\Big], \forall s_{t} \in \mathcal{S}
% % \end{smalleralign}
% A similar equation holds for $Q_\pi(s_t,a_t)$. 

Given these basic RL definitions and equations, many RL algorithms fall in the scope of generalized policy iteration (GPI). GPI is an iterative process over policy evaluation and policy improvement.

\textbf{Policy Evaluation}. The target of the policy evaluation process is to estimate state value function $V_\pi(s)$ or state-action value function $Q_\pi(s, a)$ for a given policy $\pi$. For simplicity, we only utilize $V_\pi(s)$ in the following illustration. Two common value function estimation methods are the Monte Carlo (MC) estimate and the Temporal Difference (TD) estimate \citep{sutton1988learning}. Starting from the definition of $V_\pi(s_t)$, the Monte Carlo (MC) estimate uses sampling over complete trajectories to calculate an unbiased estimate: $V_\pi(s_t) \approx \frac{1}{K} \sum_{n=1}^{K} \left[\sum_{i=t}^\infty \gamma^{i-t} r(s_i^n, a_i^n)\right]$
where we average the cumulative rewards over multiple full paths starting from $s_t$. The Temporal Difference (TD) estimate, on the other hand, builds on the temporal relationship between states without requiring complete trajectories. It estimates $V_\pi(s_t)$ using the immediate reward and the estimated value of the next state: $V_\pi(s_t) \approx \frac{1}{K} \sum_{n=1}^{K} \left[r(s_t, a_t^n) + \gamma V_\pi(s_{t+1}^n)\right]$. This approach effectively uses a "bootstrap" by depending on the estimated value of the next state to approximate $V_\pi(s_t)$.

\textbf{Policy Improvement}. The policy improvement process aims to update and improve policy according to the result from the policy evaluation process. Specifically, it replaces the old policy $\pi_{\text{old}}$ with the new one $\pi_{\text{new}}$ to increase the expected return: $V_{\pi_{\text {new }}}(s_t) \geq V_{\pi_{\text {old }}}(s_t)$. In the environment with small, discrete action spaces, such improvements can be achieved by greedily choosing the action that maximizes $Q_{\pi_{\text {old}}}(s_t, a)$: 
% \begin{smalleralign}
%     \pi_{\text {new}}(\cdot \mid s_t)=\underset{\bar{\pi}(\cdot \mid s_t) \in \mathcal{P}(\mathcal{A})}{\arg \max } \mathbb{E}_{a \sim \bar{\pi}}\left[Q_{\pi_{\text {old}}}(s_t, a)\right],
% \end{smalleralign}
\begin{smalleralign}
\label{equ:policy_improvement}
    \pi_{\text {new}}(\cdot \mid s_t)=\underset{\bar{\pi}(\cdot \mid s_t) \in \mathcal{P}(\mathcal{A})}{\arg \max } \mathbb{E}_{a \sim \bar{\pi}}\left[Q_{\pi_{\text {old}}}(s_t, a)\right],
\end{smalleralign}
Another improvement method involves applying policy gradient ascent \citep{sutton1999policy}. It parameterizes the policy $\pi_{\theta}$ with $\theta$ and we can derive the analytical policy gradient: 
\begin{smalleralign}
\nabla_\theta V_{\pi_\theta}\left(s_t\right) = \mathbb{E}_{a \sim \pi_{\theta}}\left[\nabla_\theta\log{\pi_{\theta}(a|s_t)} Q_{\pi_{\theta}}(s_t, a)\right],
\label{equ:pg}
\end{smalleralign}
By choosing a small step-size $\alpha > 0$ to conduct gradient ascent: $\theta_{\text {new }}=\theta+\left.\alpha \nabla_\theta V_{\pi_{\theta}}(s_t)\right|_{\theta=\theta_{\text {old }}}$, we can guarantee the policy improvement: $V_{\pi_{\text {new }}}(s_t) \geq V_{\pi_{\text {old }}}(s_t)$.

\section{Motivation and Methodology}
\label{sec:nlrl}
\begin{figure}[b]
    \centering
    \vspace{-15pt}\includegraphics[width=0.95\linewidth]{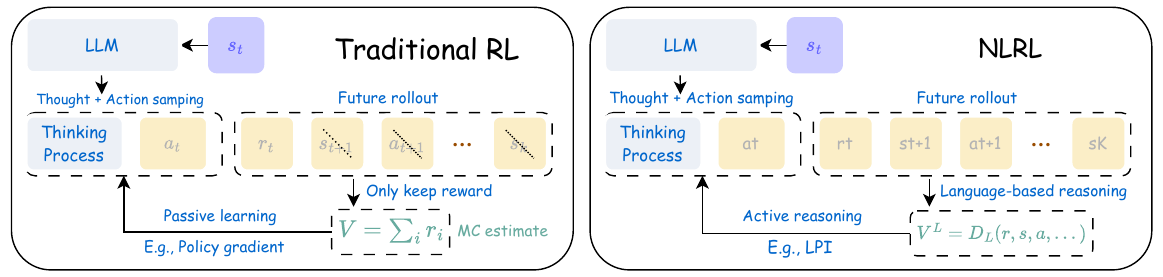}
    \caption{Comparison between NLRL and traditional RL in agentic LLM}
    \vspace{-15pt}
    \label{fig:diff}
\end{figure}
\subsection{Discussion: policy gradient is a bit 'passive' because of scalar $Q_{\pi_{\theta}}(s_t, a)$}\label{sec:discussion}
Policy gradient (\Cref{equ:pg}) is widely used in RL training, especially in recent LLM post-training trends. RL algorithms like PPO \citep{schulman2017proximal} leads to significant success in reasoning \citep{jaech2024openai,guo2025deepseek} and agent \citep{jin2025search, wang2025ragen}. These successes, however, often rely heavily on the pre-training model's prior. For example, smaller models may fail with RL training \citep{guo2025deepseek}; the capacity for test-time reasoning "reflection" may not emerge from RL, but already existed in the pre-training model \citep{liu2025understanding, shah2025rethinking}. Moreover, \citep{wang2025ragen} highlights that directly employing RL to optimize Chain-of-Thought (CoT) enhanced actions in multi-turn agentic environments can lead to CoT degradation.

The underlying cause of these challenges, we argue, stems from the inherent passive nature of policy gradient. Policy gradient algorithms improve policies by incrementally adjusting sampled action probabilities based on the scalar $Q_{\pi_{\theta}}(s_t, a)$. Effective policy improvement, therefore, becomes contingent on the agent's ability to sample high-quality reasoning chains and actions. Since $Q_{\pi_{\theta}}(s_t, a)$ only provides scalar feedback without semantic information, the agent has to passively 'wait' for beneficial actions to arise through chance, rather than actively leveraging reasoning to anticipate and strategically guide its choices. This inherent reliance on effective sampling explains why a strong pre-training model is a prerequisite for successful RL: it increases the likelihood of encountering high-reward trajectories, providing a more fertile ground for subsequent policy refinement.

The left block in \Cref{fig:diff} shows an illustration of RAGEN \citep{wang2025ragen} that trains CoT-based LLM agent with RL. RAGEN samples thinking and action from the LLM, calculate the value by taking MC estimate from future rollout, then update the LLM with policy gradient. Such MC estimate presents an extremely aggressive compression of future rollouts: it averages over the entire distribution of future outcomes, and discarding all concrete state, action or reward information, let alone any analysis. This brings in passivity: CoT may degrade if the LLM prior cannot generate diverse, beneficial thinking/action. An example of such degradation will be presented in \Cref{fig:ragen_vs_nlrl}.

\subsection{Language value function brings in active and deliberative learning}
\label{sec:lvf}
To overcome these limitations, we argue for a different approach to compressing future rollouts. Instead of collapsing the future into a single number, we propose \textbf{Language Value Function}, that leverages the expressive power of natural language, processed by the underlying LLM, to create a richer, more informative summary. This approach offers several potential advantages. First, it utilizes the LLM's inherent language understanding, reasoning and summarization capabilities, potentially drawing upon its pre-trained knowledge to generate high-quality, semantically meaningful descriptions of future possibilities. Second, it retains significantly more information than a scalar value – capturing aspects of the strategy, reasoning steps, potential pitfalls, or key events – without the impracticality of storing or processing entire raw trajectory distributions. This strikes a balance, achieving useful compression without excessive information loss.

The language state value function $V^{L}_\pi$ and language state-action value function $Q^{L}_\pi$ are defined as:
\begin{equation}
\label{equ:lvf}
\small
Q^{L}_\pi(s_t,a_t)=D_{L}\left(\mathrm{\left(s,a,r\right)}_{t+1: \infty}\sim P_\pi \mid s_t, a_t\right), V^{L}_\pi(s_{t}) = D_{L}\left(\mathrm{a_{t}, \left(s,a\right)}_{t+1: \infty}\sim P_\pi \mid s_t\right)
\end{equation}
where $D_{L}$ is a language descriptor that can transform future trajectory distribution ${(s,a,r)}_{t+1: \infty}\sim P_\pi$ into its corresponding language description.

\begin{figure}[t]
    \centering
    \vspace{-15pt}\includegraphics[width=\linewidth]{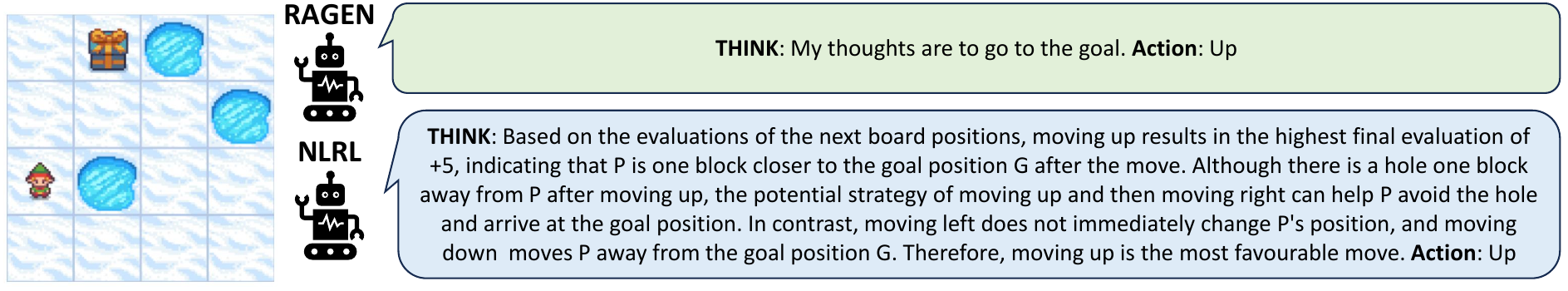}
    \caption{Comparing after-train reasoning on FrozenLake between RAGEN \citep{wang2025ragen} and NLRL (Ours). Traditional RL results in a meaningless CoT, while NLRL keeps informative CoT and active reasoning.}
    \vspace{-15pt}
    \label{fig:ragen_vs_nlrl}
\end{figure}

The right block in \Cref{fig:diff} encapsulates the features of LVF in the agentic LLM example. The LVF utilizes the LLM to perform language-based reasoning on the future rollout, generating an evaluation that captures nuanced details and offers an informative evaluation of potential outcomes. This semantically rich feedback facilitates active reasoning and strategic analysis for policy adjustment—a stark contrast to passive learning. We argue that this active reasoning path can be viewed as the agent's deliberative thinking process over task's structure. \Cref{fig:ragen_vs_nlrl} shows an example of after-train policy CoT comparison between traditional value and LVF guidance. The next section will detail how language policy improvement (LPI) can be one way to implement such active reasoning process.

\subsection{Natural language reinforcement learning}
% \section{Practical NLRL Implementation with LLMs}
% \label{sec:practice}
% Section \ref{sec:nlrl}  outlines the core philosophy of NLRL, which involves translating key concepts from RL into their human natural language counterparts. To practically implement these key concepts, a model capable of understanding, processing, and generating language is essential. Large Language Models (LLMs), trained on vast corpora of human knowledge, emerge as a natural choice to emulate human behaviour and implement language-based RL components. In the following sections, we demonstrate how LLMs can function in various roles within the NLRL framework.
% \subsection{Components and procedures}
In addition to LVF, we take one-step further: verbalizing all RL components presented in \Cref{sec:preliminary}, resulting in the NLRL framework. This includes language policy, language policy evaluation and improvement. Next, we demonstrate how LLMs can function in various roles.

\ding{172} \textbf{LLMs as language policy ($\pi_{L}$).} Many works adopted LLMs as the decision-making agent \citep{wang2023voyager, feng2023chessgpt, christianos2023pangu, yao2022react} with Chain-of-thought process \citep{wei2022cot0}. We represent the policy as $\pi_{L}(a, c|s) = \pi_{L}(c|s)\pi_{L}(a|c,s)$, including thought process $\pi_{L}(c|s)$, and the final action probability $\pi(a|c,s)$.

\ding{173} \textbf{LLMs as language value function ($Q^{L},V^{L}$).} As mentioned in \cref{sec:lvf}, we leverage LLMs, to serve as the language value function $Q^{L}, V^{L}$. This exactly corresponds to what (multi-modal) LLMs are capable of -- they are designed to take in the features from the task state, such as low-dimension statistics, text, or images, and output the corresponding language understanding. By further prompting or fine-tuning over evaluation dataset, LLMs can generate language assessment.
\begin{figure*}[t]
\vspace{-10pt}
    \centering
    \includegraphics[width=0.9\linewidth]{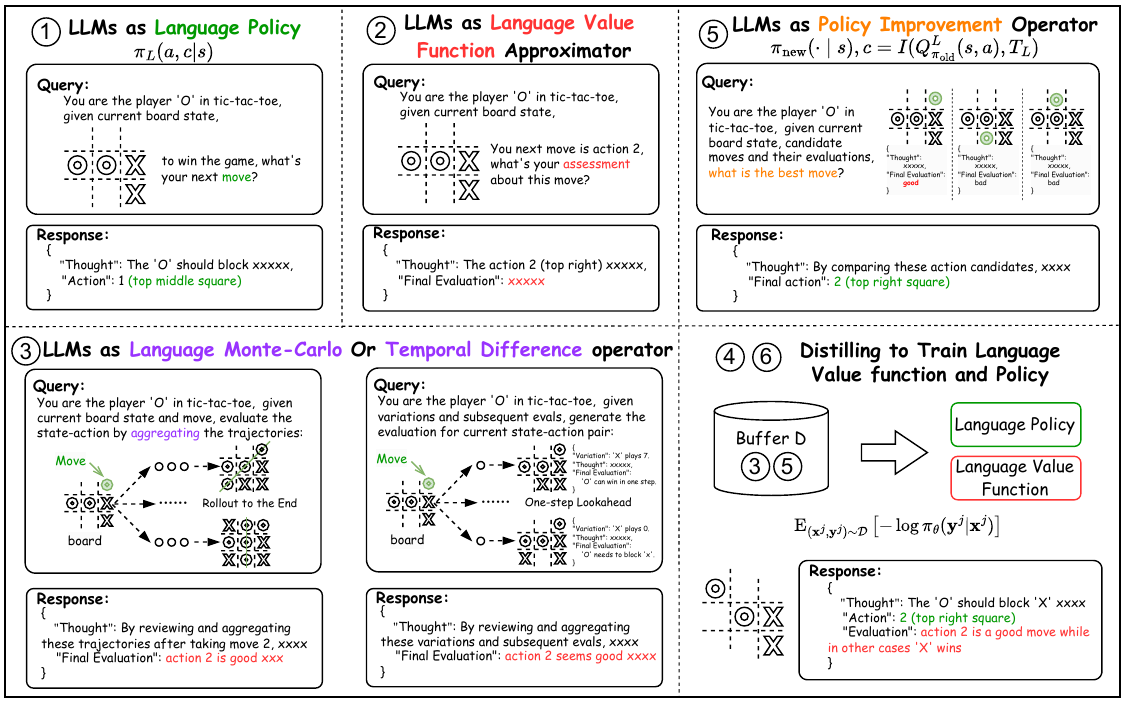}
    \caption{Practical pipeline for implementing NLRL in the Tic-tac-toe game. LLMs can serve as the language policy \ding{172}, the language-based value function approximator \ding{173}, the language Monte Carlo or Temporal Difference operator \ding{174}, and the language policy improvement operator \ding{176}. By distilling (\ding{175}, \ding{177}) the improved evaluations from \ding{173} and the enhanced actions from \ding{176}, the NLRL agent can iteratively refine its language policy and evaluation capabilities.}
    \label{fig:nlrl_implementation}
    \vspace{-10pt}
\end{figure*}

\ding{174} \textbf{LLMs as language Monte Carlo and TD operator ($G_{1}, G_{2}$)}. One core issue for language value function is: how can we generate language value estimate to train the language value function? Inspired by traditional policy evaluation methods, we propose to verbalize traditional MC and TD operations, resulting in language MC or TD estimate. Check Appendix~\ref{apx:analogy} for more about how language MC/TD corresponds to the traditional MC/TD. 

Language MC estimate is conducted over text rollouts (i.e. $K_{MC}$ full trajectories $\{a_{t}, \left(s,a\right)_{t+1: \infty}\}$) given the policy $\pi_L$. Specifically, we have $G_{1}$ that summarizes and extract the core information from different trajectories, generating language MC estimate:

\begin{equation}
\label{equ:mc_estimate}
    V^{L}_\pi(s_{t})\approx G_{1}\left(\left\{\mathrm{a_{t}^n, \left(s,a\right)}_{t+1: \infty}^{n}\right\}_{n=1}^{K_{MC}}\right),
\end{equation}

Language TD is over $K_{TD}$ few-step transitions. By drawing an analogy to Equ.~\ref{equ:one-step-bellman}, we have:
\begin{equation}
\label{eqa:td_estimate}
    V_\pi^{L}(s_t)\approx G_1\Big(\big\{G_2\big(d(s_t,a_t^n,r(s_t,a_t^n),s_{t+1}^n),V_\pi^L(s_{t+1}^n)\big)\big\}_{n=1}^{K_{TD}} \Big), \forall s_{t} \in \mathcal{S},
\end{equation}
where $d\left(a_t,r\left(s_t,a_t\right),s_{t+1})\right)$ depicts the language description of intermediate transition, while $G_{1}$ and $G_{2}$ serves as two language aggregation functions. Looking back to Equ.~\ref{equ:one-step-bellman}, $G_{2}$ mimics the add the summation operation `$+$', aggregating information from intermediate transition's description $d$ and future evaluation $V_\pi^L(s_{t+1})$, while $G_1$ serves the role of the expectation operator $\mathbb{E}$, aggregating information accross different $(a_{t},s_{t+1})$ pairs. By proper promoting, LLMs can serve as such language aggregator $G_{1}, G_{2}$ to fuse language information \citep{zhang2023benchmarking, xu2023large}.

\ding{175} \textbf{Training LVF approximator with language value estimation.} \ding{174} provides an unsupervised and scalable way to generate language evaluation data through environment interaction, which can be leveraged to train our LVF approximator in \ding{173}. This corresponds to traditional critic training for value-based RL (e.g, DQN \citep{mnih2015human}) or actor-critic algorithms (e.g, PPO \citep{schulman2017proximal}), but happens in natural language representation space. The only difference is the training objective -- the scalar-based value function regresses the value mean with mean squared error while LVF tries to imitate the language aggregation result with supervised-finetuning's cross-entropy loss.

\ding{176} \textbf{LLMs as policy improvement operator ($I$).} As mentioned in \cref{sec:lvf}, LVFs offer potentials to deliberative learning. Inspired by policy improvement (Equ.~\ref{equ:policy_improvement} and policy gradient) in traditional RL, we propose language policy improvement ($I$) to achieve as one way for deliberative learning. Specifically, for a given state $s$, we prompt the LLMs with several action candidates and their corresponding language evaluations $Q^L$, by which we can obtain the improved action with a chain-of-thought process analyzing different actions' evaluation and proposing better action. With CoT and prior knowledge about the world, LLMs are good at determining the most promising action $\pi_{\text {new}}(\cdot \mid s)$ by reasoning on different language evaluation results $Q_{\pi_{\text{old}}}^{L}(s, a)$.

\ding{177} \textbf{Training language policy with improved policy} Similar to \ding{175}, we can train our language policy LLM in \ding{172} by supervised-finetuning over the improved chain-of-thought based language policy data from \ding{176}. This corresponds to the policy training in traditional RL.
% However, NLRL relies on the more stable supervised loss instead of the policy gradient loss.

We also provide example visualizations in Fig~\ref{fig:nlrl_implementation} to illustrate these components and procedures. Building upon these components and procedures, we can create various NLRL applications. Here we illustrate three use cases, though many more possibilities exist beyond these examples.

\subsection{Language GPI boosts LLM's critic and policy by prompting (\ding{172}, \ding{173}, \ding{174}, \ding{176})}
\label{sec:use_case_1}
Our first case utilizes language GPI to enhance LLM's critic and policy solely through prompting, which can be particularly beneficial for improving proprietary models such as GPT-4 \citep{OpenAI2023GPT4TR} or Gemini \citep{team2023gemini}. Specifically, we first combine \ding{172}, \ding{173}, \ding{174} to build a language policy evaluation pipeline. Take language TD  shown in Equ.~\ref{eqa:td_estimate} as an example. We prompt LLMs to (1) evaluate the subsequent state's value $V_\pi^L(s_{t+1})$ (\ding{173}) (2) serve as the TD operator $G_1, G_2$ (\ding{174}). By performing a one-step look-ahead with $a_{t}\sim\pi$ (\ding{172}) and leveraging $G_1, G_2$ to aggregate information from intermediate transition $d$ alongside the subsequent evaluation $V_\pi^L(s_{t+1})$, language TD can generate new and improved language evaluation, $V_{\pi}^{L}(s_{t})_{\text{new}}$. Building upon this new evaluation, LLM-based policy improvement operator $I$ (\ding{176}) converts the evaluation into a better policy $\pi_{\text{new}}$. Refer to Algorithm \ref{algo:language_gpi} for detailed procedures.

\subsection{Training natural language value function for a given policy (\ding{173}, \ding{174}, \ding{175})}
\label{sec:use_case_2}
Our second case aims to train an LLM critic capable of evaluating any given state with natural language explanations, similar to a chess annotator who provides insightful commentary on boards, moves, and strategies. For example, we can build an iterative language TD pipeline by combining \ding{173}, \ding{174}, and \ding{175}. First, we leverage a tunable LLM $A$ and prompt it to become a language value function approximator \ding{173}. Combined with the look-ahead transitions $(a_t, s_{t+1})\sim P_{\pi}$ by taking rollouts with policy $\pi$, and the subsequent state evaluation $V_\pi^L(s_{t+1})$ generated by $A$, we prompt LLM $B$ for the language TD estimate (\ding{174}), similar to Sec~\ref{sec:use_case_1}. The model $A$ is further finetuned by such language TD estimates (\ding{175}) and will be plugged back for $V_\pi^L(s_{t+1})_{\text{new}}$ in a new iteration. Iteratively, we can obtain the final converged natural language value function. Check Algorithm \ref{alg:nl_language_td} for more details.

\subsection{Natural Language Actor-Critic Learning (\ding{172}, \ding{173}, \ding{174}, \ding{175}, \ding{176}, \ding{177})}
\label{sec:use_case_3}
Last, we combine all these procedures and build the full natural language actor-critic pipeline. Similar to traditional actor-critic \citep{barto1983neuronlike, sutton1999policy}, our natural language actor-critic framework simultaneously learns both a language policy and a language critic through environment interactions. For each iteration, we first use language policy \ding{172} to take rollouts in the environment. With new trajectories, we update our language value model \ding{173} by language MC or TD \ding{174} and train it with supervised-finetuning loss (\ding{175}). For language policy improvement, we query our updated language value to evaluate action candidates for states from the rollout trajectories. Further LLM-based improvement operator \ding{176} provides improved CoT data to train our language policy (\ding{177}). Check Algorithm \ref{alg:nl_actor_critic} for more details.

\section{Experiments}
\label{sec:exp}
\subsection{Language GPI by prompting (Sec~\ref{sec:use_case_1})}
\label{exp:use_case_1}

Our first experiments explore Sec~\ref{sec:use_case_1} and leverage language GPI to improve LLM capability with pure prompting. Specifically, we choose the maze games of LMRL~\citep{abdulhai2023lmrl}, aiming to validate that \textbf{Language TD Estimate} and \textbf{Language Policy Improvement} can benefit the evaluation and further improve policy. We use the original settings of LMRL Gym Maze, where the agent is required to navigate to a goal position in a ``Double T'' or ``Medium'' maze. We consider the fully-observable setting, where the agent's observation (described by text) includes the agent's current position in the maze, the agent's action history, the walls' position around the agent (if any), and the goal position. The action space is discrete, including moving up / down / right / left. We evaluate the performance on 30 different initial positions, each with 3 random seeds. Refer to Appendix~\ref{app:maze} for more details.

For language TD estimate in Equ.~\ref{eqa:td_estimate}, we prompt \textit{gpt-4o-mini-2024-07-18} as the language aggregator $G_1$, language state-action value aggregator $G_2$, and language state value function $V_{\pi}^{L}$ respectively. Specifically, given an environment state, for each candidate action, we use a fully random policy to rollout $N$ steps into the future and use the language state value function $V_{\pi}^{L}$ to evaluate the look-ahead state. For each state we repeat this process for $K$ times and use $G_2$ to aggregate into a state-action value estimation. The same GPI-4o-mini is leveraged as the policy improvement operator $I$.

We compare Language GPI with a few different baselines, including prompt-based language policy $\pi_{L}$ (\ding{172}), as well as prompt-based language value function $Q^L(s,a)$ + language policy improvement $I$. As shown in Table~\ref{tab:language_td}, Language TD produces better performance than the prompt-based language value in both mazes, and more variations \& look ahead steps can have even better results. Language policy improvement can only benefit stably for language-TD-enhanced language value (\ding{173} and \ding{174}).
\begin{table*}[t]
\vspace{-10pt}
\caption{Language GPI results with ablations on look-ahead steps $N$ and variations number $K$.}
\small
\centering
\begin{tabular}{c|c|c}
\toprule
\centering
  \textbf{Avg Reward} & \textbf{Double T Maze} & \textbf{Medium Maze} \\\midrule 
 \textit{Language policy $\pi^L(s)$ (\ding{172})} &$-27.29\pm4.43$ & $-27.05\pm5.27$ \\
 \textit{Language value function $Q^L(s,a)$ + improvement $I$ (\ding{173},\ding{176})} & $-18.33\pm6.11$ & $-33.57\pm14.41$ \\
 \textit{Language GPI (1 variation, 3 look ahead steps) (\ding{173}, \ding{174}, \ding{176})} & $-17.85\pm3.68$ & $-20.85\pm7.59$ \\
 \textit{Language GPI (4 variations, 1 look ahead steps) (\ding{173}, \ding{174}, \ding{176})} & $-17.48\pm4.53$ & $ -12.65\pm4.72$ \\
 \textit{Language GPI (4 variations, 3 look ahead steps) (\ding{173}, \ding{174}, \ding{176})} & $-12.74\pm4.47$ & $-15.09\pm4.44$   \\
 \textit{Language GPI (6 variations, 3 look ahead steps) (\ding{173}, \ding{174}, \ding{176})} & $-12.15\pm2.96$ & $-15.44\pm4.97$ \\
 \textit{Language GPI (8 variations, 3 look ahead steps) (\ding{173}, \ding{174}, \ding{176})} & $\mathbf{-11.19\pm2.86}$ & $\mathbf{-12.23\pm4.49}$  \\ \bottomrule
\end{tabular}
\vspace{-10pt}
\label{tab:language_td}
\end{table*}
\subsection{Training Natural Language Value Function with language TD (Sec~\ref{sec:use_case_2})}
\label{exp:use_case_2}

Our second experiment is an implementation of Sec~\ref{sec:use_case_2}. Specifically, we aim to train a language value function with language TD in the 5x5 breakthrough board game (Fig~\ref{Fig:5x5_breakthrough}). Breakthrough is a turn-based board game for two players, white and black. Each player has a few lines of pawns. Black moves first and in each turn, the player chooses one piece to move one square forward (straight or diagonal) onto empty squares or capture diagonally. The goal is to break through the opponent's line and reach their home row. As mentioned in Sec~\ref{sec:use_case_2}, this value function can serve as a reliable board evaluator/annotator. However, evaluating board states in this game is challenging for LLMs. First, the 5x5 breakthrough variant is niche, leading to sparse pre-training data on board evaluations. Most LLMs have minimal knowledge of it. Second, despite the small board size, its state space complexity can reach $10^8$ \citep{saffidine2012solving}, making natural language assessments significantly more difficult. Refer to Appendix~\ref{app:breakthrough} for more experimental details and result visualizations.

\begin{figure*}[t]
    \vspace{-0pt}
    \centering
    \includegraphics[width=0.75\linewidth]{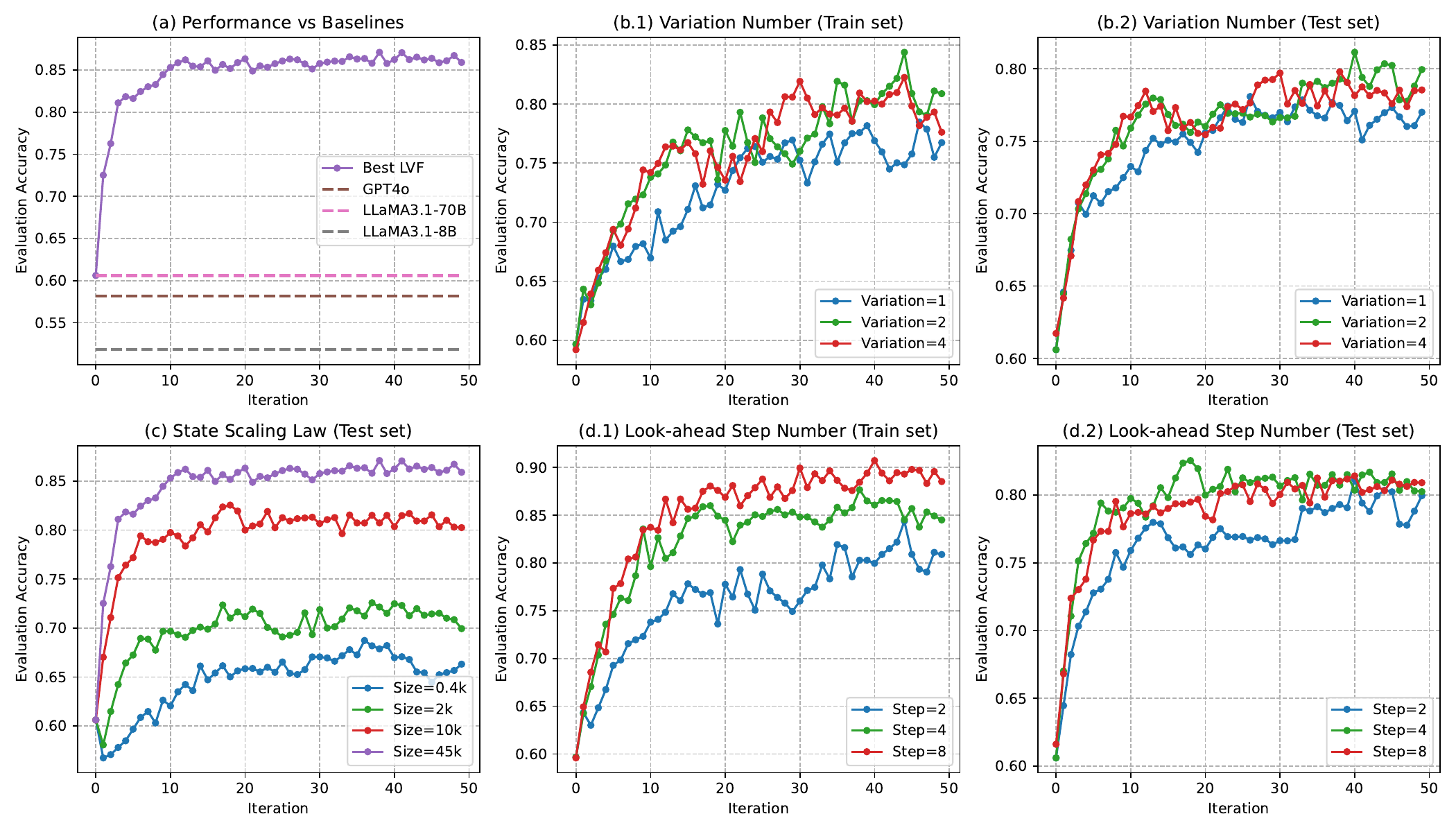}
    \vspace{-5pt}
    \caption{Breakthrough experiment results. (a) Performance comparison with baselines. (b,d) Ablation study over look-ahead step number and variation number. (c) Results for state scaling law.}
    \vspace{-18pt}
    \label{fig:breakthrough_exp}
\end{figure*}
\subsubsection{Experimental Results}
\textbf{Experimental setups.} To train a reliable board annotator, we select a strong or near-optimal policy—Monte-Carlo tree search (MCTS) algorithm \citep{kocsis2006mcts-bandit} from OpenSpiel \citep{LanctotEtAl2019OpenSpiel}, with a high number of simulations and rollouts to ensure an extensive search. For the language TD operator ($G_1, G_2$), we prompt the large \textit{LLaMA-3.1-70B-Instruct} \citep{dubey2024llama} model since it preserves stronger aggregation and reasoning ability. The language value function $V^{L}_\pi(s)$ is trained from the small \textit{LLaMA-3.1-8B-Instruct} model. For evaluation, we adopt a coarser metric—judging which side has the advantage. For any given state, we estimate win rates using Monte Carlo simulations to the game outcome. The ground-truth label is assigned to the side with a win rate exceeding a predefined threshold.

\textbf{Performance vs Baselines.} Our fig~\ref{fig:breakthrough_exp}.a shows the evaluation accuracy comparison on $D_s^{test}$, between our best language value function's training curve and prompting-based LLMs (\textit{LLaMA-3.1-70B}, \textit{LLaMA-3.1-8B} and \textit{GPT-4o-08-06}), which clearly demonstrates the necessity of language TD training. Due to the lack of domain knowledge, all prompting-based LLMs perform poorly. The best accuracy (0.61) is only slightly higher than the random guess (0.5), while our trained language value function dominates the task and the accuracy reaches more than 0.85 on the test set $D_s^{test}$.

\textbf{Look-ahead Ablation.} Fig~\ref{fig:breakthrough_exp}.b and Fig~\ref{fig:breakthrough_exp} shows further ablation study over variation number ($m$) and look-ahead steps $k$. The test set performance is shown in Fig~\ref{fig:breakthrough_exp}.b.2 and Fig~\ref{fig:breakthrough_exp}.d.2, while Fig~\ref{fig:breakthrough_exp}.b.1 and Fig~\ref{fig:breakthrough_exp}.d.1 present a subset of in-training state evaluations. The ablation study indicates that determining the appropriate $m$ and $k$ is crucial for optimal performance. Notably, in Fig~\ref{fig:breakthrough_exp}.d.1 and Fig~\ref{fig:breakthrough_exp}.d.2, using 8 steps resulted in significantly higher in-training accuracy but slightly lower test set accuracy compared to 4 steps. This suggests that longer look-ahead steps may lead to overfitting, potentially hindering generalization to novel states.

\textbf{State Scaling Law.} We also examine how scaling the training state size affects performance in Fig~\ref{fig:breakthrough_exp}.c, with datasets sizes 0.4k, 2k, 10k, and 45k. The results demonstrate robust scalability of our algorithm, with consistent improvements in test set performance as the training set size increases.

\subsection{Natural Language Actor-Critic Learning (Sec~\ref{sec:use_case_3})}
\label{exp:nlrl_ac}

For validating Sec~\ref{sec:use_case_3}, we implement the natural language actor-critic pipeline in a model-free setting~\citep{sutton2018reinforcement} on the Tic-tac-toe game~\citep{gym_tictactoe} and FrozenLake (used in \citep{wang2025ragen}). As a supplement to language TD in Sec.~\ref{exp:use_case_1} and \ref{exp:use_case_2}, we utilize the language MC (Equ.~\ref{equ:mc_estimate}) for the language policy evaluation.

\begin{figure*}[t]
    \centering
    \includegraphics[width=\linewidth]{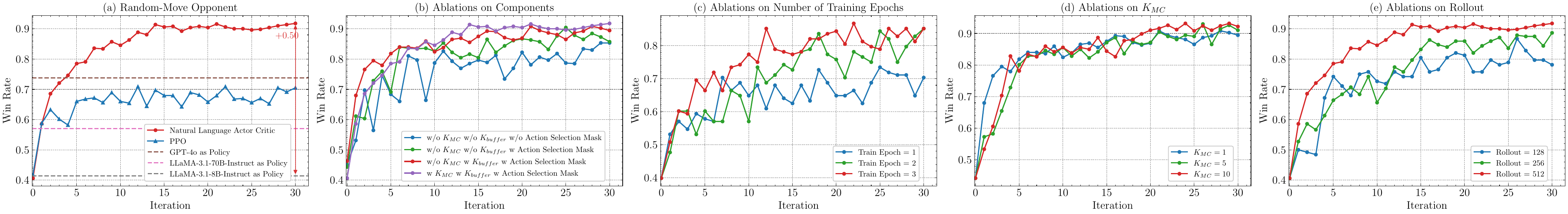}
    \vspace{-17pt}
    \caption{Natural Language Actor Critic Pipeline training results.
    (a) Training results against the Random-Move Opponent. (b) Ablation study on components ($K_{MC}$, $K_{buffer}$, and Action Selection Mask). These results demonstrate that \textbf{our proposed Natural Language Actor Critic pipeline can stably improve under stochastic dynamics}. (c) - (e) Ablation studies on number of training epochs, Monte Carlo sample size $K_{MC}$, and number of rollout trajectories.}
    \vspace{-10pt}
    \label{fig:exp-nlac-main}
\end{figure*}

% \subsubsection{Stabilizing language value function}
% \label{sec:exp-nlac-bridge}
% Despite the soundness of Sec~\ref{sec:use_case_3}, we identify training instability during initial experiments. We observed that $V_\pi^L$ can easily hallucinate when evaluating unseen state-action pairs, leading to incorrect language policy improvement. 

% We take two techniques to mitigate it. First, we enlarge the value training data size by merging the most recent $K_{buffer}$ iteration's value data buffers, which helps stabilize value training. Second, we add an action selection mask to restrict language policy improvement only to moves that are likely to be chosen by the language policy $\pi_L$ during rollout. We sample the language policy $N_{sample}$ times and build an action candidate list. The language policy improvement is conducted from the list's top-$m$ moves. Such action selection mask can help constrain policy updates within a limited range where language value function can be more reliable.

\subsubsection{Experimental Results on tic-tac-toe}
\textbf{Experiment setup.} 
Our implementation uses three language models: one \textit{LLaMA-3.1-70B-Instruct} and two \textit{LLaMA-3.1-8B-Instruct} models. The 70B model serves as the language aggregator ($G_1$) and policy improvement operator $I$. The two 8B models implement our trainable components: the language policy $\pi_L$ that generates actions through CoT reasoning, and the LVF $Q_\pi^L$ that evaluates state-action pairs. To thoroughly evaluate our approach, we test against two different types of opponents. The results against a deterministic opponent that always takes the first available action can be found in Appendix~\ref{app:tictactoe:first_move}. For our main experiments, we focus on a stochastic opponent that plays randomly, presenting a more challenging case for policy evaluation. We measure performance through game win-rate. To stabilize training and address value function instability, we implement two key techniques (More details in Appendix~\ref{sec:exp-nlac-bridge}): (1) action selection masking that restricts policy improvement to the top-$m$ moves from $N_{sample}$ samples, preventing hallucination on unseen state-action pairs; and (2) buffer merging that combines the last $K_{buffer}$ value function training sets to mitigate catastrophic forgetting. Refer to Appendix~\ref{app:tictactoe} for more experimental details.
% Our implementation uses three language models: one \textit{LLaMA-3.1-70B-Instruct} and two \textit{LLaMA-3.1-8B-Instruct} models. The 70B model is for language aggregator ($G_1$) and policy improvement operator $I$. The two 8B models implement our trainable components: language policy $\pi_L$ that generates actions through CoT reasoning, and LVF $Q_\pi^L$ that evaluates state-action pairs. To thoroughly evaluate our approach, we test against two different types of opponents. The results against a deterministic opponent that always takes the first available action can be found in Appendix~\ref{app:tictactoe:first_move}. For our main experiments, we focus on a stochastic opponent that plays randomly, presenting a more challenging case for policy evaluation. We measure performance through game win-rate. To stabilize the training, we introduce buffer merging that merge recent $K_{buffer}$ replay buffer and action selection mask technique (\cref{sec:exp-nlac-bridge}). Refer to Appendix~\ref{app:tictactoe} for more experimental details.
\paragraph{Performance vs Baselines.}
The learning curves in Fig.~\ref{fig:exp-nlac-main}(a) demonstrate our system's superiority in the random-move opponent setting. Our algorithm outperforms all other baselines (prompting-based method using \textit{GPT-4o-08-06}, \textit{LLaMA-3.1-8B/70B-Instruct} with CoT reasoning), including the PPO~\citep{schulman2017proximal} fine-tuned \textit{LLaMA-3.1-8B-Instruct} baseline. The improvement is substantial (0.4 to 0.9), demonstrating NLRL's robustness despite the stochastic dynamics. As shown in Fig.~\ref{fig:exp-nlac-main}(b), each component (Monte Carlo sampling number $K_{MC}$, buffer merging size $K_{buffer}$, and action selection mask) enhances the system's performance, with the full system achieving the highest win rate.

\textbf{Ablation Studies}
\label{exp:nlrl_hparam_abl}
We conduct comprehensive ablation studies to investigate key hyperparameters. First, training with 128 trajectories per iteration shows that increasing from 1 to 3 epochs significantly improves stability and adaptation, as shown in Fig.~\ref{fig:exp-nlac-main}(c). Second, our investigation of Monte Carlo sampling with $K_{MC}$ values of 1, 5, and 10 shows in Fig.~\ref{fig:exp-nlac-main}(d) that increased sampling with 512 trajectories leads to more stable training. Finally, Fig.~\ref{fig:exp-nlac-main}(e) shows how the number of rollout trajectories affects training, with 512 trajectories per iteration yielding the most robust learning curve. Additional ablation studies on action selection mask parameter top-$m$ and effect of experience buffer size $K_{buffer}$ can be found in Appendix~\ref{app:tictactoe:topm}, where we demonstrate that larger $m$ values enhance exploration while maintaining stable learning and $K_{buffer}=3$ helps maintain consistent performance by preserving past experiences.

\subsubsection{Experimental Results on FrozenLake}

\begin{figure*}[t]
    \vspace{-10pt}
    \centering
    \includegraphics[width=0.9\linewidth]{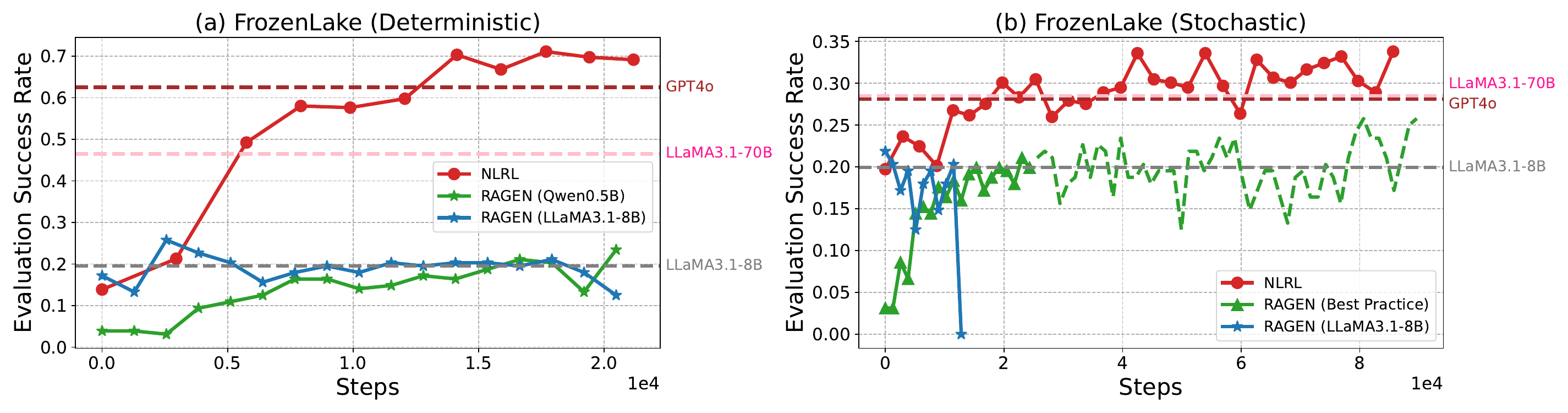}
    \vspace{-5pt}
    \caption{FrozenLake experiment results. (a) Performance comparison with baselines on FrozenLake with deterministic transition dynamics. (b) Performance comparison with baselines on FrozenLake with stochastic transition dynamics (slippery). These results demonstrate that \textbf{our proposed pipeline can more efficiently and stably improve compared to direct reinforcement learning}.}
    \vspace{-18pt}
    \label{fig:frozenlake_exp}
\end{figure*}

FrozenLake environment is a grid-based navigation task (e.g., 4×4 grids) where a player must reach a goal while avoiding holes. Following RAGEN's original implementation \cite{wang2025ragen}, we conduct similar experiments under both deterministic and stochastic transition dynamics. To test NLRL's robustness, we directly transfer our system and hyperparameters, previously tested on Tic-Tac-Toe, to FrozenLake with minimal changes. Surprisingly, even without any tuning, NLRL in in Figure \ref{fig:frozenlake_exp} demonstrates better performance on both scenarios, whereas well-tuned RAGEN suffers from inefficiency and occasional performance degradation, a phenomenon mentioned in its original paper. Notably, RAGEN’s "best practice" has slow convergence even in deterministic settings, suggesting that its inefficiency may stem not from environmental stochasticity but from its passive learning paradigm, as discussed in Section \ref{sec:discussion}. Such differences can be further visualized through the progression of reasoning trajectory (Table \ref{fig:ragen_vs_nlrl}). Refer to Appendix \ref{app:frozenlake} for more experimental details and result visualizations.

\section{Related work}
\label{sec:related_work}

\textbf{Language Model Based Autonomous Agent.} 
We are inspired by the strong emergent capabilities of LLMs in complex reasoning and planning scenarios \citep{gpt3,anil2023palm,OpenAI2023GPT4TR, feng2023alphazero, yao2022react}.
The field of language model-based autonomous agents \citep{feng2023chessgpt, christianos2023pangu, zhang2024generative} has witnessed a growing trend of leveraging LLMs for high-level planning purposes. Reflexion \citep{shinn2023reflexion} is built upon ReAct \citep{yao2022react} with self-reflection, to generate tips given online feedback. \citep{zhong2024policy} proposes to distill such reflection ability from the large model to train a small policy feedback model that can guide the policy. Their underlying ideas share similarities with language MC, while NLRL focuses on the training framework--we extract core information by sampling multiple trajectories and leverage them to train our language value function.

\textbf{Learning from Language Feedback.} Our work is also related to research on learning from language feedback. While \citep{cheng2023llf} focuses on benchmarking algorithms, we aim to propose a new algorithmic framework. Studies such as \citep{yang2023large, yuksekgonul2024textgrad, cheng2024trace} introduce an LLM-based optimization paradigm that leverages natural language to represent optimization operators like gradients and backpropagation, achieving end-to-end generative optimization. NLRL, on the other hand, represents a parallel approach, offering a generative framework specifically designed for RL optimization problems.

\textbf{LLMs as evaluation function.} Our language value function aligns with recent efforts in NLP that leverage LLMs as generative evaluators or verifiers, as seen in \citep{wang2023pandalm, li2023generative,jiang2023tigerscore, gao2024llm, zhang2024generative, mahan2024generative}, or that utilize LLMs’ self-evaluation capabilities to enhance task-solving performance \citep{putta2024agent, yao2023tot, hao2023rap}. These approaches often rely on the LLM’s self-assessment (which can be unreliable), or distill on stable supervision signals like human annotations or guidance from stronger LLMs. In contrast, NLRL’s training of a language value function offers a novel unsupervised approach.

\textbf{Interpretable Reinforcement Learning.}
Our work is also related to interpretable RL, such as methods of concept-based explanations for sequential decision-making tasks. \citep{sreedharan2020bridging, hayes2017improving, das2023state2explanation} formulates concept-based explanations upon state preconditions and action costs.

\section{Conclusion}
\label{sec:conclusion}
In this work, we propose a new paradigm NLRL, which reformulates the RL process with natural language representation. We believe NLRL opens a new door for understanding and implementing RL algorithms in natural language space, improving agent's effectiveness, efficiency and enabling active learning. NLRL also serves as a potential way for generating high-quality language synthetic data for policy and critic, which could be crucial in training more advanced language models and agents. We leave further discussion and limitation in Appendix~\ref{apx:limitation_and_future_work}.
\newpage
% \section*{Impact Statement}

% This work introduces Natural Language Reinforcement Learning (NLRL), which aims to advance both reinforcement learning and natural language processing by reformulating RL concepts in natural language space. Beyond the technical contributions, we identify several potential broader impacts:
% First, by making RL systems more interpretable through natural language representations, NLRL could enhance transparency and accountability in AI systems. This increased interpretability may facilitate better human oversight and understanding of AI decision-making processes, particularly important as these systems are deployed in real-world applications.
% Second, NLRL's capability to learn from environment feedback without human annotation could reduce the reliance on expensive human labeling. While this efficiency gain is beneficial, it may impact the employment landscape for data annotators and raise questions about the role of human expertise in AI training.
% Third, as NLRL enables more autonomous learning in new domains, careful consideration should be given to potential misuse, especially in sensitive applications. The framework's ability to learn from environment interaction alone necessitates robust safety measures and ethical guidelines for deployment.
% Finally, the computational resources required for training language models in NLRL may have environmental implications through increased energy consumption. Future work should consider optimizing the framework's efficiency to minimize its environmental footprint.

% \nocite{langley00}
\bibliographystyle{abbrvnat}
\bibliography{icml/icml2025}

%%%%%%%%%%%%%%%%%%%%%%%%%%%%%%%%%%%%%%%%%%%%%%%%%%%%%%%%%%%%%%%%%%%%%%%%%%%%%%%
%%%%%%%%%%%%%%%%%%%%%%%%%%%%%%%%%%%%%%%%%%%%%%%%%%%%%%%%%%%%%%%%%%%%%%%%%%%%%%%
% APPENDIX
%%%%%%%%%%%%%%%%%%%%%%%%%%%%%%%%%%%%%%%%%%%%%%%%%%%%%%%%%%%%%%%%%%%%%%%%%%%%%%%
%%%%%%%%%%%%%%%%%%%%%%%%%%%%%%%%%%%%%%%%%%%%%%%%%%%%%%%%%%%%%%%%%%%%%%%%%%%%%%%
\clearpage
\appendix
\onecolumn

% \vspace{3ex}
% \begin{center}
% {\Large\textbf{Supplementary Material}}
% \end{center}
% \vspace{2ex}
\section*{Contents}
\startcontents
\printcontents{}{0}[3]{}
\newpage

\section{Detailed Related work}
\label{sec:apx_related_work}

\textbf{Language Model Based Autonomous Agent.} 
Inspired by the strong emergent capabilities of LLMs in complex reasoning and planning scenarios \citep{gpt3,wei2022emergent,anil2023palm,OpenAI2023GPT4TR, feng2023alphazero}.
The field of language model-based autonomous agents \citep{feng2023chessgpt, christianos2023pangu, zhang2024generative} has witnessed a growing trend of leveraging LLMs for high-level planning purposes. 
For text agents, ReAct\citep{yao2022react} leverages chain-of-thought planning about the intermediate goals and task-specific generation via few-shot prompting. Reflexion \citep{shinn2023reflexion} is built upon ReAct \citep{yao2022react} with self-reflection, named verbal reinforcement learning, to generate tips given online feedback. Such tips strengthen agent's capability through in-context learning. \citep{zhong2024policy} proposes to distill such reflection ability from the large model to train a small policy feedback model that can guide the policy. Their underlying ideas share similarities with NLRL's language Monte Carlo estimate, while NLRL formulates this process more formally and incorporates it into training--we extract core information/concepts by sampling multiple trajectories and leverage them to train our natural language value function. Another line of work like \citep{zhang2023benchmarking, xu2023large} leverages Large Language Models to summarize and extract information to help fuse intermediate changes. They are related to the "aggregation" idea of language value function training.

\textbf{Interpretable Reinforcement Learning.}
One of the major purposes of interpretable RL is to automatically seek explanations from non-AI experts. For instance, methods of concept-based explanations for sequential decision-making tasks. \citep{ji2023spatial} provide a concept-based explanation for 3D action recognition CovNets by clustering learned human interpretable features. \citep{sreedharan2020bridging} formulates concept-based explanations upon state preconditions and action costs, representing any factual statement a user associates with a given state. Similarly, \citep{hayes2017improving} uses logical formulas to summarize policies. 
Additionally, \citep{das2023state2explanation} trains a joint embedding model for state-action pairs and concept-based explanations.

\textbf{Learning from Language Feedback.} Our work is also related to research on learning from language feedback. While \citep{cheng2023llf} focuses on benchmarking algorithms, we aim to propose a new algorithmic framework. Studies such as \citep{yang2023large, yuksekgonul2024textgrad, cheng2024trace} introduce an LLM-based optimization paradigm that leverages natural language to represent optimization operators like gradients and backpropagation, achieving end-to-end generative optimization. NLRL, on the other hand, represents a parallel approach, offering a generative framework specifically designed for RL optimization problems.

\textbf{LLMs as evaluation function.} Our language value function aligns with recent efforts in NLP that leverage LLMs as generative evaluators or verifiers, as seen in \citep{wang2023pandalm, li2023generative,jiang2023tigerscore, gao2024llm, zhang2024generative, mahan2024generative}, or that utilize LLMs’ self-evaluation capabilities to enhance task-solving performance \citep{putta2024agent, yao2023tot, hao2023rap}. These approaches often rely on the LLM’s self-assessment (which can be unreliable), or distill on stable supervision signals like human annotations or guidance from stronger LLMs. In contrast, NLRL’s training of a language value function offers a novel unsupervised approach: we can train a generative evaluator solely by aggregating language feedback from the environment, without distilling any strong model--experimental results in Sec.~\ref{exp:use_case_2} show that our language value function's performance is much better than any LLMs we utilize during the training.

\textbf{Successor Features.} Successor features \citep{dayan1993improving, barreto2017successor} aim to learn a high-dimensional value function representation, where each component captures the expected future occurrence of all state representations under a fixed policy. As shown in \citep{barreto2017successor}, successor features help decouple the environment dynamics from the rewards, facilitating transfer learning in RL. The language value function in NLRL is conceptually similar to successor features since it also represents state values in a high-dimensional space and satisfies the language Bellman equation.

% \newpage
\section{Algorithms}\label{apx:more_algorithms}
In this section, we provide the full pseudo-code for three cases mentioned in Sec.~\ref{sec:use_case_1}, ~\ref{sec:use_case_2}, and ~\ref{sec:use_case_3}.
\subsection{Language GPI by prompting}
\begin{algorithm}[htb] 
\caption{Language GPI by Prompting} 
\label{algo:language_gpi}\begin{algorithmic}[1] \REQUIRE Initial language policy $\pi$, language value function $V_\pi^L$, LLM implementing operators $G_1$, $G_2$, policy improvement operator $I$, number of lookahead steps $N$, number of variations $K$ \FOR{each state $s_t$} \STATE // Language TD estimation \FOR{$i=1$ to $K$} \STATE Sample action $a_t^{(i)}$ \hfill (\ding{172}) \STATE Simulate $N$ steps ahead to get trajectory $\tau^{(i)}$ \STATE Prompt LLM to evaluate $V_\pi^L(s_{t+N}^{(i)})$ \hfill (\ding{173}) \STATE Generate intermediate description $d^{(i)} = d(a_t^{(i)}, r_t^{(i)}, s_{t+1}^{(i)})$ \STATE Compute value estimate $V^{(i)} = G_2(d^{(i)}, V_\pi^L(s_{t+N}^{(i)}))$ \hfill (\ding{174}) \ENDFOR \STATE Aggregate value estimates $V_\pi^L(s_t) = G_1(\{V^{(i)}\}_{i=1}^{K})$ \hfill (\ding{174}) \STATE // Policy improvement \STATE Use $V_{\pi}^L(s_t)$ and prompt LLM as operator $I$ to select the optimal action $a_t$ \hfill (\ding{176}) \ENDFOR \end{algorithmic} \end{algorithm}
\subsection{Train natural language language value function with language TD}
\begin{algorithm}[htb]
\caption{Train natural language language value function with language TD}
\label{alg:nl_language_td}
\begin{algorithmic}[1]
\REQUIRE Pre-defined policy $\pi$, LLM-based language value model $V^L_\pi$, language descriptor $d$, language aggregator $G_1$, $G_2$, state buffer $D_s$, number of iterations $T$, number of trajectories $N$, number of MC trajectories $K$
\STATE // Collect few-step rollout data using fixed policy for TD data buffer
\STATE Initialize TD data buffer $\mathcal{B} \leftarrow \{\}$
\FOR{state $s_t$ in state buffer $D_s$}
    \STATE Starting from $s_t$, run policy $\pi$ for a few $l$-step look-ahead rollouts and collect $K$ distinct variations: $V_{s_t}=\{(s_{t+i}, a_{t+i}, r_{t+i}, s_{t+i+1})_{i=0}^{l-1}\}_{K}$ 
    \STATE $\mathcal{B} \leftarrow \mathcal{B} \cup V_{s_t}$
\ENDFOR
\STATE // Language TD learning
\FOR{iteration $t=1$ to $T$}
    \STATE // Language TD estimate \ding{173}, \ding{174}
    \STATE $\mathcal{D}_V \leftarrow \{\}$ \COMMENT{Value training data}
    \FOR{each look-ahead data $(s_t, V_{s_t})$ in $\mathcal{B}$}
        \STATE For each variation $n$, generate the final state evaluation $V_{\pi}^L(s_{t+l}^{n})$ with $V_{\pi}^L$, and utilize $d$ to describe intermediate transitions $d_n=d(s_{t}, a_{t}^{n}, r_{t}^{n}, s_{t+1}^{n},..., s_{t+l-1}^{n}, a_{t+l-1}^{n}, r_{t+l-1}^{n})$
        \STATE $V_\pi^{L}(s_t)_{new} \leftarrow G_1\Big(\big\{G_2\big(d_n,V_\pi^L(s_{t+1}^n)\big)\big\}_{n=1}^K \Big)$ \COMMENT{l-step, k-variation Language TD}
        \STATE $\mathcal{D}_V \leftarrow \mathcal{D}_V \cup V_\pi^{L}(s_t)_{new}$
    \ENDFOR
    
    \STATE // Update value function \ding{175}
    \STATE Train $V^L_\pi$ using language TD estimate dataset $\mathcal{D}_V$ with supervised loss
\ENDFOR
\end{algorithmic}
\end{algorithm}

\newpage

\subsection{Natural language actor critic}
\begin{algorithm}[!htbp]
\caption{Natural Language Actor-Critic Learning}
\label{alg:nl_actor_critic}
\begin{algorithmic}[1]
\REQUIRE Initial language policy model $\pi_L$, language value model $V^L_\pi$, large language model for evaluation $G_1$/$G_1$,$G_2$, improvement operator $I$
\REQUIRE Number of iterations $T$, number of trajectories $N$, number of MC trajectories $K$
\STATE Initialize replay buffer history $\mathcal{B} \leftarrow \{\}$
\FOR{iteration $t=1$ to $T$}
    \STATE // Collect trajectories using language policy \ding{172}
    \STATE $\tau \leftarrow \{\}$ \COMMENT{Initialize trajectory buffer}
    \FOR{$i=1$ to $N$}
        \STATE Run policy $\pi_L$ to collect trajectory: $\tau_i \leftarrow \{(s_j, a_j, r_j, s_{j+1})_{j=0}^{H_i}\}$
        \STATE $\tau \leftarrow \tau \cup \{\tau_i\}$
    \ENDFOR

    \STATE // Language value estimation \ding{173}, \ding{174}
    \STATE $\mathcal{D}_V \leftarrow \{\}$ \COMMENT{Value training data}
    \FOR{each state-action pair $(s, a)$ in $\tau$}
        \IF{using MC estimate}
            \STATE Sample $K_{MC}$ trajectories starting from $(s, a)$: $\{\tau_k\}_{k=1}^{K_{MC}}$
            \STATE $v \leftarrow G_1(\{\tau_k\}_{k=1}^{K_{MC}})$ \COMMENT{Language MC estimate}
        \ELSIF{using TD estimate}
            \STATE Sample $K_{TD}$ transitions starting from $(s, a)$: $\{(s_k, a_k, r_k, s'_k)\}_{k=1}^{K_{TD}}$
            \STATE $v \leftarrow G_1(\{G_2(d(s_k,a_k,r_k,s'_k), V^L_\pi(s'_k))\}_{k=1}^{K_{TD}})$ \COMMENT{Language TD estimate}
        \ENDIF
        \STATE $\mathcal{D}_V \leftarrow \mathcal{D}_V \cup \{(s, a, v)\}$
    \ENDFOR
    
    \STATE // Update value function \ding{175}
    \STATE $\mathcal{B} \leftarrow \mathcal{B} \cup \{\mathcal{D}_V\}$ \COMMENT{Add to buffer history}
    \STATE Train $V^L_\pi$ on merged data from last $K_{buffer}$ buffers in $\mathcal{B}$ with supervised loss

    \STATE // Language policy improvement \ding{176}
    \STATE $\mathcal{D}_\pi \leftarrow \{\}$ \COMMENT{Policy training data}
    \FOR{each state $s$ in $\tau$}
        \STATE Sample actions $\{a_i\}_{i=1}^{N_{sample}}$ from $\pi_L(s)$
        \STATE Select top-$m$ actions $\mathcal{A}_k$ based on sampling frequency
        \STATE Query value estimates: $Q^L \leftarrow \{V^L_\pi(s,a) | a \in \mathcal{A}_k\}$
        \STATE $\pi_{new}, c \leftarrow I(Q^L)$ \COMMENT{Language improvement operator}
        \STATE $\mathcal{D}_\pi \leftarrow \mathcal{D}_\pi \cup \{(s, \pi_{new}, c)\}$
    \ENDFOR
    
    \STATE // Update policy \ding{177}
    \STATE Train $\pi_L$ on $\mathcal{D}_\pi$ with supervised loss
\ENDFOR
\end{algorithmic}
\end{algorithm}

\newpage
\section{Experimental details}
\label{apx:exp}
\subsection{Maze Experiment}
\subsubsection{Experiment Details}

\begin{wrapfigure}{r}
{0.355\textwidth}
    \centering
\vspace{-12pt}
\includegraphics[width=0.195\textwidth]{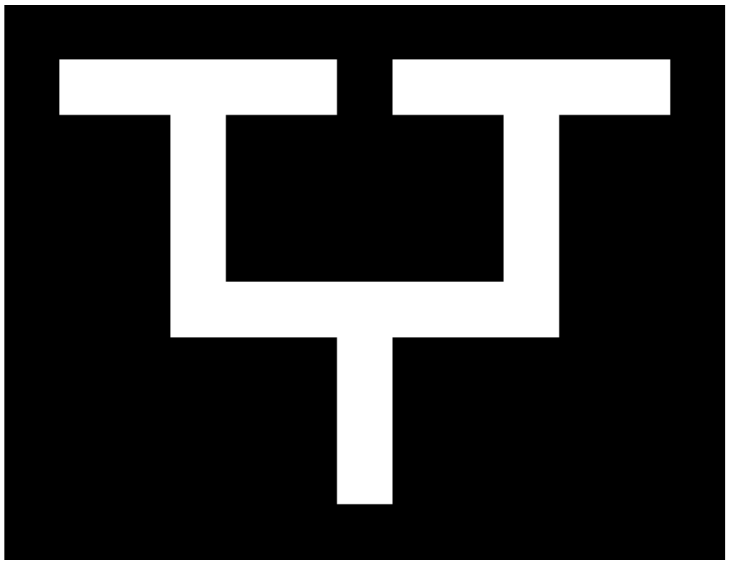}
\includegraphics[width=0.15\textwidth]{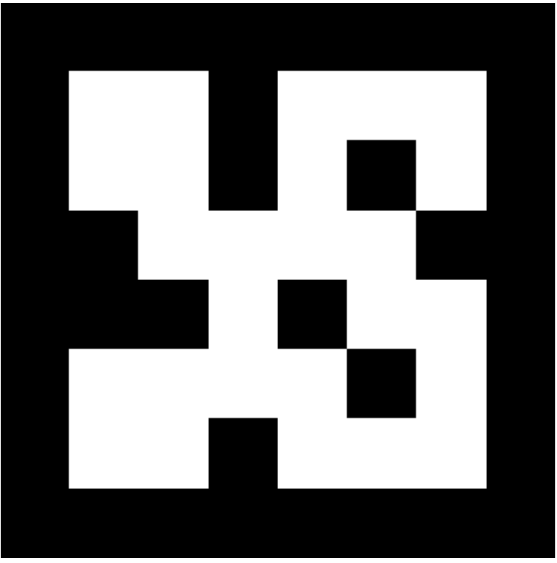}
\vspace{-13pt}
    \caption{The ``double T'' maze and the ``medium'' maze. At the beginning of each test, the agent randomly respawns in white space.}
    \vspace{-13pt}
    \label{fig:maze_env}
\end{wrapfigure}

Our Maze experiment is conducted using LMRL Gym~\citep{abdulhai2023lmrl}. We use the original codebase for the environment. We consider the fully-observable setting, where the agent's observation (described by text) includes the agent's current position in the maze, the agent's action history, the walls' position around the agent (if any), and the goal position. The action space is discrete, including moving up / down / right / left. We evaluate the performance on 30 different initial positions, each with 3 random seeds.

We prompt \textit{gpt-4o-mini-2024-07-18} (temperature $=0.7$) as the language aggregator $G_1$, language state-action value aggregator $G_2$, and language state value function $V_{\pi}^{L}$ respectively. 
\subsubsection{Language GPI Visualization}
\label{app:maze}
We present the language GPI's prompt and response visualization for language state value function $V_{\pi}^{L}$ and language state-action value aggregator $G_2$ in the LMRL Maze experiment.
\begin{mdframed}[style=mystyle, frametitle={Maze prompts}]
\addcontentsline{toc}{paragraph}{LMRL Gym Maze prompts}
\small\ttfamily
\begin{longtable}{|p{0.15\textwidth}|p{0.8\textwidth}|}
\hline
\textbf{Role} & \textbf{Content} \\ \hline
SYSTEM & You are an expert maze solver. You only respond in json.\\ \hline
USER ($V_{\pi}^{L}$) & You are playing a simple maze game. An agent is moving \\ &in the maze and the objective is to reach the goal in as few steps as possible. 

The possible actions are "move up", "move down", "move left", "move right".

You are a highly skilled evaluator in this game and is expected to function similar to state value function in reinforcement learning. At each step I will provide you with the move history of the agent (from old to new), including where the goal is, your current position, the walls that surround the agent. Your output evaluation should be a json array that includes the following concepts:

- "thoughts": Let's think step by step. Generate your detailed thought process and evaluation of the agent's position and the distance towards the goal.

- "final evaluation": Concisely judge how good the agent's current position is compared to all the possible positions in the maze, in terms of reaching the goal.

Let's start a new game. Now, please give your evaluation of the current state given the move history of the agent:

The goal is at position 8, 6. Your current position is at position 5, 7. There are walls above you, below you.\\ &move up \\ &The goal is at position 8, 6. Your current position is at position 5, 7. There are walls above you, below you.\\ &move down \\ &The goal is at position 8, 6. Your current position is at position 5, 7. There are walls above you, below you.\\ &move right\\ &The goal is at position 8, 6. Your current position is at position 5, 8. There are walls above you, below you.\\ &move right\\ &The goal is at position 8, 6. Your current position is at position 5, 9. There are walls to your right, below you. \\ \hline

ASSISTANT & 'thoughts': ["The goal is located at (8, 6) and the agent's current position is (5, 9).", 'To evaluate the distance to the goal, I will calculate the Manhattan distance, which is the sum of the absolute differences of the coordinates: |8 - 5| + |6 - 9| = 3 + 3 = 6.', "The agent's current position (5, 9) has walls directly below (6, 9) and to the right (5, 10). This limits movement options.", 'The possible actions for the agent are to move left to (5, 8) or move up to (4, 9). Moving down or right is not possible due to walls.', 'Moving left to (5, 8) would bring the agent back to a previous position, while moving up to (4, 9) opens potential new pathways.', 'Overall, the agent is relatively far from the goal, with limited movement options due to surrounding walls.'], \\ &'final evaluation': "The agent's current position (5, 9) is not ideal as it is 6 steps away from the goal and has limited movement options. Moving up to (4, 9) is the best immediate option to explore new paths." \\ \hline
\end{longtable}

\begin{longtable}{|p{0.15\textwidth}|p{0.8\textwidth}|}
\hline
\textbf{Role} & \textbf{Content} \\ \hline
SYSTEM & You are an expert maze solver. You only respond in json.\\ \hline

USER $(G_{2})$ & You are playing a simple maze game. An agent is moving in the maze and the objective is to reach the goal in as few steps as possible. 

The possible actions are "move up", "move down", "move left", "move right".

You are a highly skilled evaluator in this game, particularly adept at making accurate assessments through look-ahead of the current maze position after taking the given action. At each step I will provide you with the move history of the agent (from old to new), including where the goal is, your current position, the walls that surround the agent, the action that the agent is going to take, *along with several key variations of trajectory pieces after taking this action (and the corresponding natural language evaluations of the trajectory pieces)*. 
Your task is to understand these look-ahead information and summarize, derive non-trivial analysis and understanding the *the agent's position after taking the action*. Your output evaluation should be a json array with the following *two* concepts:
\\ &"thoughts": Let's think step by step. Summarize the look-ahead information of the variations after taking action move down.
\\ &"final evaluation": Now Concisely judge how good the chosen action is, in terms of reaching the goal.

Now, please give your evaluation given action move down, the *current environment state*:

The goal is at position 8, 6. Your current position is at position 5, 7. There are walls above you, below you

and the look-ahead information of different variations after taking action move down: 

\textbf{AGGREGATION OF VALUE ESTIMATES FOR ALL VARIATIONS} \\ \hline

ASSISTANT & 'thoughts': ["The goal is located at (8, 6), which is three rows down and one column to the left of the current position (5, 7).",    "The agent attempts to move up from (5, 7) but cannot because there is a wall above.",   "Each variation shows the agent getting stuck in a loop of ineffective moves after trying to move up, consistently returning to (5, 7) or moving to (5, 9), which is still not closer to the goal.",   "The walls surrounding the agent restrict movement, making it impossible to progress toward the goal efficiently.",  "In all variations, the agent finds itself either returning to previous positions or moving to areas blocked by walls.",    "The agent\'s progression towards the goal becomes increasingly limited as it cannot move up, down, or directly toward the goal from its current position."], \\ &'final evaluation': "\{    "current position": "(5, 7)",   "goal position": "(8, 6)",   "manhattan distance": 6,     "possible moves": ["move down", "move left"],    "blocked moves": ["move up", "move right"],   "evaluation": "The chosen action \'move up\' is ineffective as it does not change the agent\'s position, and the surrounding walls create a dead end, making it impossible to reach the goal efficiently."\}" \\ \hline
\end{longtable}
\end{mdframed}

\subsubsection{The effect of Language TD on the Value function}
Below we show a comparison of the Language state-action values with/without Language TD. Comparing the two Language State-action Value, we can see that by analyzing various potential outcomes, Language TD identifies both immediate and future benefits or drawbacks of actions, which is reflected in its more detailed thoughts and final evaluation. The inclusion of varied trajectory information allows Language TD to provide a more balanced and informed assessment, mitigating over-reliance on immediate next-step evaluations.
\begin{mdframed}[style=mystyle, frametitle={Language State-action Value (Language TD)}]
\addcontentsline{toc}{paragraph}{Language State-action Value (Language TD)}
\small\ttfamily
\begin{spverbatim}
For action "move down", {"thoughts": ["The agent's current position is (5, 7) with the goal at (8, 6).", "After attempting to move down, the agent remains at (5, 7) due to walls above and below.", "In Variation 1, the agent eventually moves to (6, 6), which is one step away from the goal at (8, 6).", "Variation 2 shows that moving down leads to blocked paths, necessitating lateral movement, but ultimately keeps the agent away from the goal.", "Variation 3 allows the agent to successfully reach (7, 6), positioning it just one step away from the goal.", "Variation 4 also confirms that moving down to (6, 6) is effective, placing the agent two steps away from the goal.", "Overall, the evaluations indicate that while the agent faces wall constraints, moving down leads to favorable outcomes that progress toward the goal."], "final_evaluation": {"current_position": "(5, 7)", "distance_to_goal": 3, "proposed_action": "move down", "evaluation": "Although the move down is initially blocked, subsequent evaluations indicate that this action can lead to optimal positions in the following moves. The agent can reach the goal within two more steps if it navigates correctly."}}
\end{spverbatim}
\end{mdframed}
\begin{mdframed}[style=mystyle, frametitle={Language State-action Value (without Language TD)}]
\addcontentsline{toc}{paragraph}{Language State-action Value (without Language TD)}
\small\ttfamily
\begin{spverbatim}
For action "move down", {
  "thoughts": [
    "The agent's current position is (5, 7) and the goal is at (8, 6).",
    "The possible action is to move down to (6, 7).",
    "There are walls above (4, 7) and below (6, 7), which means the agent can only move down if it is not blocked.",
    "Moving down to (6, 7) is valid since there is no wall at that position.",
    "After moving down, the new position will be (6, 7).",
    "The distance to the goal can be calculated as follows: the horizontal distance is |8 - 6| = 2 and the vertical distance is |6 - 7| = 1.",
    "The total distance to the goal after moving down would be 2 + 1 = 3."
  ], "final_evaluation": {
    "current_position": "(6, 7)",
    "distance_to_goal": 3,
    "evaluation": "The agent's position after moving down is better than the previous one, as it moves closer to the goal. There are no walls blocking further movement towards the goal from this position."
  }
}

\end{spverbatim}
\end{mdframed}

\subsection{Breakthrough Experiment}
\label{app:breakthrough}
\subsubsection{Experiment Setup and Methods}
\textbf{Text-Breakthrough.} As mentioned in the text-based MDP, we textualize the game, including board, pieces, moves, positions, and information about capturing or termination.

\textbf{Policy $\pi$ in $V_\pi^L(s)$.} Our first task is to determine the $\pi$ in $V_\pi^L(s)$ since $V_\pi^L(s)$ measures the value for a specific policy, according to Equ~\ref{equ:lvf}. To train a reliable board annotator, we select a strong or near-optimal policy—analogous to the way we place greater trust in a grandmaster's annotations. In our experiment, our $\pi$ is the Monte-Carlo tree search (MCTS) algorithm \citep{kocsis2006mcts-bandit} from OpenSpiel \citep{LanctotEtAl2019OpenSpiel}, with a high number of simulations and rollouts to ensure an extensive search.

\textbf{Building State Dataset $s\sim P(s)$.} Since $V_\pi^L(s)$ is over state $s$, our second step is to build a state dataset $D_s$--the distribution $P(s)$ that our $V_\pi^L$ works on. To ensure that $V_\pi^L$ can assess positions across diverse levels of gameplay, we build a mixed-state dataset by collecting rollouts between different MCTs configurations. Then we can easily split $D_s$ to build training state set $D_s^{train}$ and test state set $D_s^{test}$.
% Specifically, we maintain a pool of MCTS policies, where each policy variant is characterized by a distinct number of simulations and rollouts. The mixed state dataset is then created by pairwise combining policies from this pool, running rollouts for each policy pair, and merging the resulting data from all combinations.

\textbf{Building TD training dataset.} The TD training dataset is built on $D_s^{train}$by conducting look-ahead expansion with rollout policy $\pi$. For each state $s_t$, we take a few times $l$-step look-ahead rollout and deduplicate to $K$ distinct variations: $\{(s_{t+i}, a_{t+i}, r_{t+i}, s_{t+i+1})_{i=0}^{l-1}\}_{K}$.

\textbf{Models.} For the language TD operator ($G_1, G_2$), we prompt the large \textit{LLaMA-3.1-70B-Instruct} \citep{dubey2024llama} model since it preserves stronger aggregation and reasoning ability. The language value function $V^{L}_\pi(s)$ is trained from the small \textit{LLaMA-3.1-8B-Instruct} model.

\textbf{Evaluation.}  An automatic evaluation metric is challenging even for advanced LLMs like GPT-4o, given their limited domain knowledge. We adopt a coarser metric—judging which side has the advantage. For any given state, we estimate win rates using Monte Carlo simulations to the game outcome. The ground-truth label is assigned to the side with a win rate exceeding a predefined threshold.
\subsubsection{More Details}

\begin{wrapfigure}{r}{0.3\textwidth}
    \centering
    \vspace{-15pt}
\includegraphics[width=0.3\textwidth]{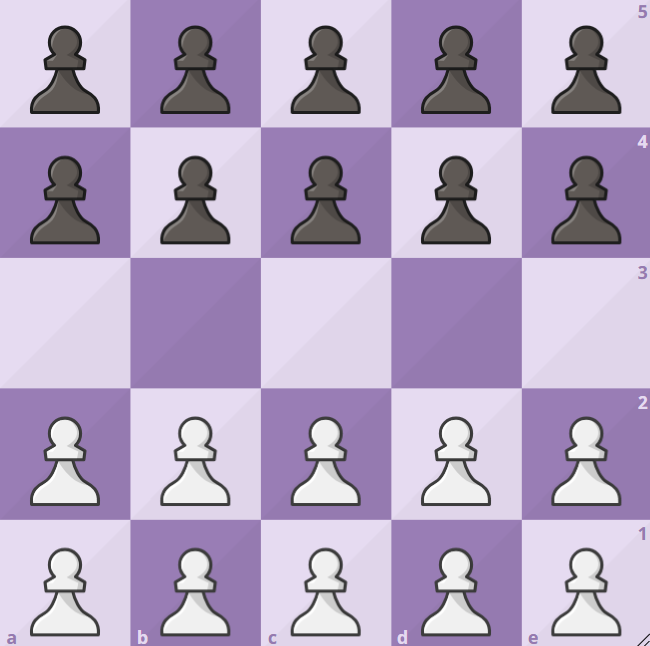}
\caption{5x5 Breakthrough.}
\vspace{-15pt}
\label{Fig:5x5_breakthrough}
\end{wrapfigure}

Our breakthrough experiment is conducted with Openspiel \citep{LanctotEtAl2019OpenSpiel} 5x5 breakthrough game and MCTS policy. We slightly modified Openspiel's codebase so the 5x5 breakthrough game supports two lines of pieces for each side.

For the policy $\pi$, as mentioned in the paper, we choose an MCTS policy with $uct_c=1$, 1000 simulations for search, and 100 rollouts for the random rollout evaluator. For the state dataset, we define the MCTS policies using a grid of configurations based on the number of simulations and rollouts. Specifically, we use the following values for these parameters:

\begin{itemize}
    \item \textbf{Simulation numbers:} \(2\), \(10\), \(100\), \(1000\)
    \item \textbf{Rollout numbers:} \(1\), \(10\), \(100\), \(1000\)
\end{itemize}

This results in \(4 \times 4 = 16\) unique MCTS policies, each characterized by a specific combination of simulation and rollout numbers. To generate the mixed state dataset, we pair every possible combination of these 16 policies (including self-pairing), leading to \(16 \times 16 = 256\) policy pairs.

For each pair of policies, we perform the same number of rollouts where both policies interact within the environment. The states encountered during these rollouts are recorded, ensuring that the dataset captures a diverse distribution of positions. By merging the data from all policy pairs, we create the final mixed state dataset \(D_s\), which is designed to represent states arising from a wide range of gameplay strategies and skill levels.

We use vLLM~\citep{kwon2023efficient} for any LLM inference used in the experiment, including language value function inference and aggregation inference.

We sample 3000 states from the state dataset for evaluation to serve as the $D_{test}^s$. When computing prediction accuracy, we only count state evaluation which can pass our rule-based parser to extract the advantageous side judgment.

Here we present the hyperparameters used in our experiment:

\textbf{Rollout}

\begin{table}[H]
\centering
\begin{tabular}{@{}ll@{}}
\toprule
\textbf{Parameter} & \textbf{Value} \\
\midrule
Parallel Environments & 192 \\
Lookahead Step & 4 \\
Lookahead Rollout Number & 4 \\
Deduplicate State & True \\
\bottomrule
\end{tabular}
\caption{Rollout Parameters}
\end{table}

\textbf{Prompting}

\begin{table}[H]
\centering
\begin{tabular}{@{}ll@{}}
\toprule
\textbf{LLM Sampling Parameter} & \textbf{Value} \\
\midrule
Temperature & 1.0 \\
Top K & 50 \\
Top P & 0.95 \\
Max Tokens & 512 \\
\bottomrule
\end{tabular}
\caption{LLM sampling parameters for prompting.}
\end{table}

\textbf{Training}

\begin{table}[H]
\centering
\begin{tabular}{@{}ll@{}}
\toprule
\textbf{Parameter} & \textbf{Value} \\
\midrule
Max Sequence Length & 1024 \\
Warmup Ratio & 0.03 \\
Learning Rate & 2e-5 \\
Learning Rate Scheduler & Constant \\
Dtype & bfloat16 \\
Per Device Train Batch Size & 4 \\
Gradient Accumulation Step & 8 \\
Training Epoch & 2 \\
Number of GPUs & 4 \\
Distributed Framework & FSDP \\
\bottomrule
\end{tabular}
\caption{Data Collection Parameters}
\end{table}

\textbf{Evaluation.} We use the same LLM sampling parameters as the prompting process.

\subsubsection{Language TD Visualization}
Here we present an example of how language TD works for query and response.

\begin{mdframed}[style=mystyle, frametitle={Language State-action Value (Before)}]
\addcontentsline{toc}{paragraph}{Language State-action Value (Before)}
\small\ttfamily
\begin{longtable}{|p{0.15\textwidth}|p{0.8\textwidth}|}
\hline
\textbf{Role} & \textbf{Content} \\ \hline
SYSTEM & Here is the rule for the Breakthrough board game:\\ 
& The game is played on an 5x5 board for 2 players (white and black), with each player starting with 10 pawns. white pawns are on the first two rows and black pawns are on the last two rows.\\
& Black moves first. In each turn, players can move one of their pieces one square forward, diagonally forward if the target square is empty. Or it can capture an opponent's piece if that square is one step diagonally forward.\\
&The game ends when one player successfully break through the opponent lines -- either move a piece to the opposite last row of the board or captures all of the opponent's pieces.\\\\
&For board representation, we use b for black pieces, w for white pieces, and . for empty squares. (1-5) and (a-e) are used to show the rows and columns index respectively.\\\\
&You are a highly skilled evaluator in this game, particularly adept at making accurate assessments through look-ahead analysis of the current board position.\\
& I will provide you with current board representation, *along with several key variations starting from this position (and their corresponding natural language evaluations of the subsequent positions)*.\\\\
&Your task is to aggregate and compare these look-ahead information, to summarize, derive non-trivial analysis about the *current board*. Your output should include the following concepts:\\
&1. *Tactical Considerations*: Any immediate threats, potential tactics, or vulnerabilities in the position.\\
&2. *Positional Evaluation*: Consideration of pawn structure, piece activity, control of key squares, and game safety.\\ &3. *Suggested Moves*: One or two strong candidate moves for the side to move, along with a brief rationale for comparing different moves.\\
&4. *Advantage*: Based on all previous rationale, determine if white or black takes advantage. Use <white> or <black> to represent the winning side.\\\\
&Your response should be informative and concise. \\ \hline
USER & "*The board you need to evaluate:* \\&\\&5bb.b.\\&4b..b.\\&3..bw.\\&2w.w..\\&1wwwww\\& abcde\\&\\& It is White's turn.\\&White pieces are at: d3, a2, c2, a1, b1, c1, d1, e1.\\&Black pieces are at: a5, b5, d5, a4, d4, c3.\\&\\&\\&Here are the look-ahead variations from the current board position:\\&*Key Variations and Subsequent Evaluation:*:\\&\\&*Variation 1:* \\&Description of variation's move sequence:\\&The action sequence is: d3e4,d5e4*,a2b3,a4b3*.\\&Move 1:White moves piece from d3 (Column d, Row 3) to e4 (Column e, Row 4).\\&Move 2:Black moves piece from d5 (Column d, Row 5) to e4 (Column e, Row 4), capturing White piece.\\&Move 3:White moves piece from a2 (Column a, Row 2) to b3 (Column b, Row 3).\\&Move 4:Black moves piece from a4 (Column a, Row 4) to b3 (Column b, Row 3), capturing White piece.\\&\\&Subsequent position evaluation:\\&The subsequent board is: \\&\\&5bb...\\&4...bb\\&3.bb..\\&2..w..\\&1wwwww\\& abcde\\&\\& It is White's turn.\\&White pieces are at: c2, a1, b1, c1, d1, e1.\\&Black pieces are at: a5, b5, d4, e4, b3, c3.\\&\\&\\&The evaluation of this subsequent board is: \\&\\&**Current Board Analysis:**\\&\\&**Tactical Considerations:**\\&White's most significant tactical consideration is the potential to capture Black's pieces on d4 and e4, which would gain a significant advantage. White's pieces are well-positioned to launch a decisive attack. Black's pieces on a5 and b5 are somewhat isolated and vulnerable to potential attacks.\\&\\&**Positional Evaluation:**\\&The current position slightly favors White due to their piece activity, control of key squares, and potential to create a strong pawn center. White's pawns on a1 and d1 provide potential support for the central pawns. Black's pawns are somewhat isolated, but they still have a chance to reorganize.\\&\\&**Suggested Moves:**\\&Based on the look-ahead analysis, two strong candidate moves for White are:\\&\\&1. **e1-d2**: This move captures Black's potential piece on d2 (in case Black plays d4-d2) and creates a strong initiative.\\&2. **c2-b3**: This move attacks Black's position and prepares to break through to the opposite side, gaining a strong initiative.\\&\\&**Advantage:**\\&<white>\\&\\&Overall, White has a slight advantage in this position, with multiple ways to break through Black's lines and gain a significant advantage. However, Black's position is not yet critical, and they can still reorganize their pieces to defend. White's advantage is mainly due to their piece activity, control of key squares, and potential to create a strong pawn center.\\&\\&*Variation 2:* \\&Description of variation's move sequence:\\&The action sequence is: c1b2,c3d2,e1d2*,d4e3.\\&Move 1:White moves piece from c1 (Column c, Row 1) to b2 (Column b, Row 2).\\&Move 2:Black moves piece from c3 (Column c, Row 3) to d2 (Column d, Row 2).\\&Move 3:White moves piece from e1 (Column e, Row 1) to d2 (Column d, Row 2), capturing Black piece.\\&Move 4:Black moves piece from d4 (Column d, Row 4) to e3 (Column e, Row 3).\\&\\&Subsequent position evaluation:\\&The subsequent board is: \\&\\&5bb.b.\\&4b....\\&3...wb\\&2wwww.\\&1ww.w.\\& abcde\\&\\& It is White's turn.\\&White pieces are at: d3, a2, b2, c2, d2, a1, b1, d1.\\&Black pieces are at: a5, b5, d5, a4, e3.\\&\\&\\&The evaluation of this subsequent board is: \\&\\&**Tactical Considerations:**\\&The current position presents a strong opportunity for White to create a breakthrough on the 'd' file, as demonstrated by the two variations. Black's pawn structure is somewhat disjointed, making it difficult to respond to White's advances on the 'd' and 'e' files.\\&\\&**Positional Evaluation:**\\&White's pawn structure is more compact and cohesive, controlling key central squares. The pawns on 'a' and 'b' files provide additional support for the potential breakthrough on 'd' and 'e'. Black's pieces on 'a5' and 'd5' are somewhat isolated, making it difficult to coordinate a defense.\\&\\&**Suggested Moves:**\\&Considering the look-ahead variations, the strongest candidate move for White is **d3d4**. This move puts pressure on Black's position and sets up the potential breakthrough on the 'd' and 'e' files. The rationale for this move is that it creates an immediate threat and forces Black to respond, potentially weakening their position.\\&\\&Another possible move could be **d2d3**, but it is less aggressive and does not put as much pressure on Black.\\&\\&**Advantage:**\\&Based on the analysis, the advantage is **<white>**. The two variations demonstrate a clear path to victory for White, and Black's position appears to be under significant pressure.\\&\\&Please provide your analysis and understanding of the current board position based on the provided look-ahead information.\\&Your response should be informative and concise. \\ \hline

ASSISTANT & "**Current Board Analysis:**\\&\\&**Tactical Considerations:**\\&The primary tactical consideration for White is to utilize the pawns on the 'd' file to create a breakthrough, potentially capturing Black's pieces on d4 and e4. Additionally, White should focus on controlling the key squares and putting pressure on Black's somewhat isolated pieces on a5 and b5. Black needs to reorganize their pawns and pieces to defend against White's central pawn advance.\\&\\&**Positional Evaluation:**\\&The position slightly favors White due to their more compact and cohesive pawn structure, piece activity, and control of key central squares. White's pawns on the 'a' and 'b' files provide additional support for the potential breakthrough on the 'd' and 'e' files. Black's pawns are somewhat isolated, making it challenging to coordinate a defense.\\&\\&**Suggested Moves:**\\&Based on the look-ahead analysis, two strong candidate moves for White are:\\&\\&1. **d3d4**: This move puts pressure on Black's position, sets up a potential breakthrough on the 'd' and 'e' files, and forces Black to respond.\\&2. **e1-d2**: This move captures Black's potential piece on d2 and creates a strong initiative.\\&\\&**Advantage:**\\&<white>\\&\\&Overall, White has a slight advantage in this position, with multiple ways to break through Black's lines and gain a significant advantage. The two variations demonstrate a clear path to victory for White, and Black's position appears to be under significant pressure. White's advantage is mainly due to their piece activity, control of key squares, and potential to create a strong pawn center.\\&\\&The provided look-ahead information suggests that White should prioritize controlling the central squares, putting pressure on Black's position, and utilizing the pawns on the 'd' file to create a breakthrough."
 \\ \hline
\end{longtable}
\end{mdframed}
\subsubsection{Comparison of language value function}
Here we present another visualization comparison between the initial \textit{LLaMA-3.1-8B-Instruct} Model and our trained model. The evaluation is conducted on the board 5..b.b
4.bb..
3.w...
2.w...
1ww..w
 abcde, where white takes large advantage in this place. By our language TD training, the language value function can accurately evaluate this board and generate insightful annotation, largely decreasing the hallucination.
\begin{mdframed}[style=mystyle, frametitle={Language State Value (Before)}]
\addcontentsline{toc}{paragraph}{Language State Value (Before)}
\textbf{Board: 5..b.b
4.bb..
3.w...
2.w...
1ww..w
 abcde, white turn}
\small\ttfamily
\begin{spverbatim}
 Based on the current board position, here's the evaluation for White's turn:

*Tactical Considerations:*

- Black has an opportunity to attack White's pawn at e1 with a pawn from e5 or c5, which might put White in a difficult position.
- White's pawn at b3 and b2 is a target for Black's pawns at b4 and c4.
- White needs to focus on safely advancing their pawns and creating a barrier for Black's pieces.

*Positional Evaluation:*

- White has a relatively weak pawn structure, with some pawns on the first row.
- The center is somewhat controlled by White, but Black has a solid presence on the fifth row.
- Black has a strategic advantage in pawn structure, but White has the opportunity to influence the game with proper planning.

*Suggested Moves:*

1. Move the pawn from b2 to b3. This move has the goal of preparing to defend and possibly create a barrier against Black's pieces.
2. Move the pawn from a1 to a2. This move aims to create more space and put pressure on Black's pieces, which will make it harder for them to maneuver. 

*Advantage:* 
Based on the current board position, <black> is favored.
\end{spverbatim}
\end{mdframed}

\begin{mdframed}[style=mystyle, frametitle={Language State Value (After training)}]
\addcontentsline{toc}{paragraph}{Language State Value (After training)}
\textbf{Board: 5..b.b
4.bb..
3.w...
2.w...
1ww..w
 abcde, white turn}
\small\ttfamily
\begin{spverbatim}
**Analysis and Evaluation of the Current Board Position**

**Tactical Considerations:**
The key variation shows an immediate threat of White breaking through Black's lines. White's piece on b3 is strategically positioned to capture Black's piece on c4, which creates an opportunity for a decisive push.

**Positional Evaluation:**
White has a significant advantage in terms of piece activity and control of key squares. With pieces on the third and second rows, White is poised to push forward and break through Black's lines. Black's pieces are concentrated on the fifth and fourth rows, which limits their mobility and ability to respond to White's attack.

**Suggested Moves:**
Two strong candidate moves for White are:

1. b3-c4*: This move captures Black's piece and creates a weakness in Black's position, allowing White to potentially break through.
2. b2-c3: This move develops another White piece and prepares to support the potential breakthrough on the c-file.

However, given the look-ahead variations, the most promising move is b3-c4*, as it leads to a terminal state where White wins.

**Advantage:**
Based on the analysis, White has a decisive advantage. The ability to capture Black's piece on c4 and break through Black's lines gives White a strong initiative and a clear path to victory. Therefore, <white> is the winning side.
\end{spverbatim}
\end{mdframed}

\subsection{Tic-tac-toe Experiment}
\label{app:tictactoe}
\subsubsection{Experiment Setup and Methods}
\textbf{Text-Tic-tac-toe.} Similar to text-breakthrough, we textualize all information in the Tic-tac-toe.

\textbf{Model Architecture.} 
Our implementation uses three language models: one \textit{LLaMA-3.1-70B-Instruct} and two \textit{LLaMA-3.1-8B-Instruct} models. The 70B model is for language aggregator ($G_1$) and policy improvement operator $I$. The two 8B models implement our trainable components: language policy $\pi_L$ that generates actions through chain-of-thought reasoning, and language value function $Q_\pi^L$ that evaluates state-action pairs.
% - notably, we restrict this model to use only its general language processing and information aggregation capabilities, rather than any game-specific knowledge it may have learned during pre-training. This restriction is enforced through carefully designed prompts (see Appendix~\ref{app:prompts:tictactoe:policy-evaluation-prompt} for details)

% \textbf{Dataset Construction.} 
% For each iteration, we collect 512 complete game trajectories with $\pi_L$. For Monte Carlo value estimation, we collect $K_{MC}=5$ complete trajectories for each state-action pair. For policy improvement, $\pi_L$ proposes $N_{sample}=10$ candidate actions per state and we select the top-$m$ ($m=10$) actions from the candidate action list.

\textbf{Evaluation.}
% To thoroughly evaluate our approach, we test against two different types of opponents. The first is a deterministic opponent that always takes the first available action, while the second is a stochastic opponent that plays randomly, presenting a more challenging case for policy evaluation. We measure performance through metrics including win/loss/tie rates.
To thoroughly evaluate our approach, we test against two different types of opponents. The results against a deterministic opponent that always takes the first available action can be found in Appendix~\ref{app:tictactoe:first_move}. For our main experiments, we focus on a stochastic opponent that plays randomly, presenting a more challenging case for policy evaluation. We measure performance through metrics including win/loss/tie rates.

\textbf{Baselines.} We compare our algorithm with prompting-based method using GPT-4o-08-06, LLaMA-3.1-8B/70B-Instruct with step-by-step reasoning. We also include a traditional RL baseline by fine-tuning LLaMA-3.1-8B-Instruct with standard PPO~\citep{schulman2017proximal}.
% Experiments on Tic-tac-toe are conducted using \url{https://github.com/haje01/gym-tictactoe}~\citep{gym_tictactoe} for the environment and vLLM for efficient language model inference. We implement a parallel data collection framework using 64 environments to gather game trajectories efficiently.
\subsubsection{More details}
Our experiments are conducted on the Tic-tac-toe environment~\citep{gym_tictactoe}, a standard two-player game environment that provides a clear testbed for evaluating sequential decision-making. The implementation uses several key components: \url{https://github.com/haje01/gym-tictactoe} (MIT License) for the Tic-tac-toe environment, FSDP (Fully Sharded Data Parallel) for distributed model training, vLLM~\citep{kwon2023efficient} for efficient language model inference, and a custom parallel framework for trajectory collection. The environment is wrapped with a parallel execution layer to enable efficient batch processing of multiple games simultaneously.

For policy evaluation, we employ Monte Carlo sampling to estimate state-action values, requiring complete game trajectories until terminal states. Policy improvement utilizes a structured sampling approach with temperature-controlled exploration and frequency-based action selection.

To manage computational resources efficiently, we employ gradient checkpointing and use the FSDP strategy across our GPU cluster. This configuration allows us to train larger language models while maintaining reasonable memory requirements and training speed.

The parallel data collection framework is designed to maximize GPU utilization during training. It maintains a queue of active environments and processes state-action transitions in batches, significantly reducing the overall training time compared to sequential execution.

\textbf{Model Architecture.} 
Our implementation uses three language models: one \textit{LLaMA-3.1-70B-Instruct} and two \textit{LLaMA-3.1-8B-Instruct} models. The 70B model implements the language aggregator ($G_1$) and policy improvement operator $I$ - notably, we restrict this model to use only its general language processing and information aggregation capabilities, rather than any game-specific knowledge it may have learned during pre-training. This restriction is enforced through carefully designed prompts (see Appendix~\ref{app:prompts:tictactoe:policy-evaluation-prompt} for details). The two 8B models implement our trainable components: one serves as the language policy $\pi_L$, generating actions through chain-of-thought reasoning, while the other implements the language value function $Q_\pi^L$, evaluating state-action pairs.

\textbf{Dataset Construction.} 
For each iteration, we collect 512 complete game trajectories with $\pi_L$. For Monte Carlo value estimation, we collect $K_{MC}=5$ complete trajectories for each state-action pair. For policy improvement, $\pi_L$ proposes $N_{sample}=10$ candidate actions per state and we select the top-$m$ ($m=10$) actions from the candidate action list.

\textbf{Evaluation.}
To thoroughly evaluate our approach, we test against two different types of opponents. The first is a deterministic opponent that always takes the first available action, while the second is a stochastic opponent that plays randomly, presenting a more challenging case for policy evaluation. We measure performance through metrics including win/loss/tie rates.

\subsubsection{Stabilizing language value function}
\label{sec:exp-nlac-bridge}
Despite the soundness of Sec~\ref{sec:use_case_3}, we identify training instability resulting from the language value function, mainly from two perspectives. First, we observed that the language value function $V_\pi^L$ can easily hallucinate when evaluating unseen state-action pairs, leading to incorrect language policy improvement. To address this, we add an action selection mask to restrict language policy improvement only to moves that are likely to be chosen by the language policy $\pi_L$ during rollout. Instead of prompting LLM to select from all available moves, we sample the language policy $N_{sample}$ times and build an action candidate list. The language policy improvement is conducted from the list's top-$m$ moves. Such selection mask can help constrain policy updates within a limited range, akin to how KL divergence is used in RLHF objective function \citep{ouyang2022training}.

Second, we encountered the challenge of catastrophic forgetting for value training. Specifically, LLM can rapidly overfit to the state-action pairs encountered in the current iteration while forgetting past experiences, e.g. state-action pairs encountered only a few iterations ago. To address this, we enlarge the language value function training set by merging the last $K_{buffer}$ value function buffers, which help stabilize the training of the language value function by preserving past knowledge.

\subsubsection{Training Infrastructure}

\textbf{Data Collection Pipeline}

For each training iteration, we collect data using the following configuration:

\begin{table}[H]
\centering
\begin{tabular}{@{}ll@{}}
\toprule
\textbf{Parameter} & \textbf{Value} \\
\midrule
Parallel Environments & 64 \\
Trajectories per Iteration & 512 \\
Monte Carlo Samples ($K_{MC}$) & 5 \\
Policy Samples per State ($N_{sample}$) & 10 \\
Top-k Actions & 10 \\
\bottomrule
\end{tabular}
\caption{Data Collection Parameters}
\label{tab:data_collection_params}
\end{table}

\textbf{Model Architecture and Training}

Both policy and value networks are trained from \textit{LLaMA-3.1-8B-Instruct} using the following parameters:

\begin{table}[H]
\centering
\begin{tabular}{@{}ll@{}}
\toprule
\textbf{Hyperparameter} & \textbf{Value} \\
\midrule
Learning Rate & 1e-5 \\
Learning Rate Schedule & Constant \\
Training Epochs per Iteration & 2 \\
FSDP Configuration & Full Sharding \\
Gradient Checkpointing & Enabled \\
Batch Size & 8 \\
Max Sequence Length & 1024 \\
Training Hardware & 4 × H100 GPUs \\
\bottomrule
\end{tabular}
\caption{Model Training Hyperparameters}
\label{tab:training_params}
\end{table}

\textbf{Value Function Buffer Management}

To prevent catastrophic forgetting, we maintain experience buffers with the following configuration:
\begin{table}[H]
\centering
\begin{tabular}{@{}ll@{}}
\toprule
\textbf{Parameter} & \textbf{Value} \\
\midrule
Buffer History Length ($K_{buffer}$) & 3 iterations \\
Merging Strategy & Equal sampling \\
Buffer Content & State-action pairs with MC estimates \\
\bottomrule
\end{tabular}
\caption{Buffer Management Configuration}
\label{tab:buffer_params}
\end{table}

\subsubsection{Algorithm Implementation Details}

\textbf{Policy Evaluation Details}

For Monte Carlo evaluation, we use the following configuration:
\begin{table}[H]
\centering
\begin{tabular}{@{}ll@{}}
\toprule
\textbf{Parameter} & \textbf{Value} \\
\midrule
MC Trajectories per State-Action & 5 \\
Trajectory Completion & Full game \\
Value Aggregation & Average over returns \\
Sampling Temperature & 0.7 \\
Action Space Size & 9 positions (0-8) \\
\bottomrule
\end{tabular}
\caption{Policy Evaluation Configuration}
\label{tab:policy_eval_params}
\end{table}

\subsubsection{Visualization}
% Below we show a comparison of the Language state-action values before and after Language MC.

Below we show a comparison of the Language state-action values before and after Language MC estimation. The evaluation is conducted on the following board:

\[
\begin{array}{|c|c|c|}
\hline
\text{O} & \text{X} & 3 \\
\hline
4 & \text{O} & \text{O} \\
\hline
7 & \text{X} & 9 \\
\hline
\end{array}
\] 

Through analyzing multiple complete game trajectories, Language MC helps the value function develop a comprehensive understanding of long-term consequences of actions, rather than just immediate rewards. This is particularly evident in the enhanced evaluation quality after MC estimation, where the model provides more strategic insights and nuanced assessment of game positions.

\begin{mdframed}[style=mystyle, frametitle={Language State-action Value (Before)}]
\addcontentsline{toc}{paragraph}{Language State-action Value (Before)}
\small\ttfamily
\begin{longtable}{|p{0.15\textwidth}|p{0.8\textwidth}|}
\hline
\textbf{Role} & \textbf{Content} \\ \hline
SYSTEM & You are a player of the game of Tic Tac Toe. The game goal is to get multiple of one's own symbols in a row, either horizontally, vertically, or diagonally, before the opponent does. If all nine squares are filled and no player has three in a row, the game is a draw. \\
       & The board consists of "O", "X" and grid numbers. The grid number indicates an empty grid. You are learning how to evaluate a (board, action) pair in the Tic Tac Toe game by playing the game given the (board, action) pair and reflecting on the playing history. \\
       & The playing history depicts a game in progress with a final result. Your answer consists of two elements: \\
       & - "thought": Let's think step by step. Generate your detailed evaluation over the (board, action) pair by merely reflecting the playing history after this pair from different perspectives. You should only rely on the playing history as context and shouldn't evaluate the game with your own judgment. Your evaluation should contain the following elements: Win probability, Threat, and Potential strategies. \\
       & - "final\_evaluation": After all of your thoughts, judge which side takes advantage. 1 means you think 'O' takes advantage, -1 means you think 'X' takes advantage, and 0 means you think the game will be a draw. \\
       & You should output your answer in the JSON format. \\
       & \textbf{EXAMPLE:} \\
       & user: The (board, action) to evaluate is O's turn: \\
       & Board: \\
       & O | O | X \\
       & --------- \\
       & 4 | X | 6 \\
       & --------- \\
       & 7 | 8 | 9 \\
       & Action: The O's move is 4. \\
       & Below is the rollout sequence after this (board, action): \\
       & After O's move 4, the board position is: \\
       & O | O | X \\
       & --------- \\
       & O | X | 6 \\
       & --------- \\
       & 7 | 8 | 9 \\
       & After X's move 7, the board position is: \\
       & O | O | X \\
       & --------- \\
       & O | X | 6 \\
       & --------- \\
       & X | 8 | 9 \\
       & The game is over. X wins. X wins by occupying the positions 3, 5, 7. \\
       & assistant: \\
       & \{ \\
       & "thought": \{ \\
       & "Reflection": "It appears that the initial board position \\
       & O | O | X \\
       & --------- \\
       & 4 | X | 6 \\
       & --------- \\
       & 7 | 8 | 9 \\
       & and action 4 were not favorable for O, as X was able to block on O's moves and ultimately win the game.", \\
       & "Win probability": "The win probability for X is large, while the win probability for O is low.", \\
       & "Threat": "X has played center 5 and corner 3. X can win by playing corner 7. O was able to occupy 1, 4, 7 and create a potential winning opportunity.", \\
       & "Potential strategies": "Potential strategies for O include playing the corner 7 to block X as opposite corner and win by occupying 1, 4, 7. X could have occupied 3, 5, 7 to win the game. X has already occupied 3, 5, and there is 1 step to complete the plan." \\
       & \}, \\
       & "final\_evaluation": -0.9 \\
       & \} \\ \hline
USER & The board to evaluate is O's turn: \\
     & Board: \\
     & O | X | 3 \\
     & --------- \\
     & 4 | O | 6 \\
     & --------- \\
     & 7 | X | 9 \\
     & Action: The O's move is 3. \\
     & Below is the rollout sequence 1 after this (board, action): \\
     & After O taking action 3, the board position is: \\
     & O | X | O \\
     & --------- \\
     & 4 | O | 6 \\
     & --------- \\
     & 7 | X | 9. \\
     & After X taking action 4, the board position is: \\
     & O | X | O \\
     & --------- \\
     & X | O | 6 \\
     & --------- \\
     & 7 | X | 9. \\
     & After O taking action 7, the board position is: \\
     & O | X | O \\
     & --------- \\
     & X | O | 6 \\
     & --------- \\
     & O | X | 9. \\
     & The game is over. O wins. O wins by occupying the positions [3, 5, 7]. \\
     & Now generate your evaluation for the (board, action) pair. \\ \hline

ASSISTANT & \{ \\
          & "thought": \{ \\
          & "Reflection": "The initial board position \\
          & O | X | 3 \\
          & --------- \\
          & 4 | O | 6 \\
          & --------- \\
          & 7 | X | 9 \\
          & and action 3 were favorable for O, as O was able to occupy a strategic position and ultimately win the game.", \\
          & "Win probability": "The win probability for O is high, while the win probability for X is low.", \\
          & "Threat": "O has played center 5 and corner 3. O can win by playing corner 7. X has played center 5's opposite corner 9 but failed to block O's winning line.", \\
          & "Potential strategies": "Potential strategies for O include playing the corner 7 to win by occupying 3, 5, 7. X could have occupied 4, 6, 9 to block O's winning line but failed to do so." \\
          & \}, \\
          & "final\_evaluation": 0.9 \\
          & \} \\ \hline
\end{longtable}
\end{mdframed}

% \begin{mdframed}[style=mystyle, frametitle={Language State-action Value (Before)}]
% \addcontentsline{toc}{paragraph}{Language State-action Value (Before)}
% \small\ttfamily
% \begin{spverbatim}
% {"thought": {"Reflection": "The initial board position 
% O | X | 3
% ---------
% 4 | O | 6
% ---------
% 7 | X | 9
% and action 3 were favorable for O, as O was able to occupy a strategic position and ultimately win the game.", "Win probability": "The win probability for O is high, while the win probability for X is low.", "Threat": "O has played center 5 and corner 3. O can win by playing corner 7. X has played center 5's opposite corner 9 but failed to block O's winning line.", "Potential strategies": "Potential strategies for O include playing the corner 7 to win by occupying 3, 5, 7. X could have occupied 4, 6, 9 to block O's winning line but failed to do so."}
% "final_evaluation": 0.9}
% \end{spverbatim}
% \end{mdframed}

\begin{mdframed}[style=mystyle, frametitle={Language State-action Value (After)}]
\addcontentsline{toc}{paragraph}{Language State-action Value (After)}
\small\ttfamily
\begin{longtable}{|p{0.15\textwidth}|p{0.8\textwidth}|}
\hline
\textbf{Role} & \textbf{Content} \\ \hline
SYSTEM & You are a player of the game of Tic Tac Toe. The game goal is to get multiple of one's own symbols in a row, either horizontally, vertically, or diagonally, before the opponent does. If all nine squares are filled and no player has three in a row, the game is a draw. \\
       & The board consists of "O", "X" and grid numbers. The grid number indicates an empty grid. You are learning how to evaluate a (board, action) pair in the Tic Tac Toe game by playing the game given the (board, action) pair and reflecting on the playing history. \\
       & The playing history depicts a game in progress with a final result. Your answer consists of two elements: \\
       & - "thought": let's think step by step. Generate your detailed evaluation over the (board, action) pair by merely reflecting the playing history after this pair from different perspectives. You should only rely on the playing history as context and don't evaluate the game with your own judgment. Your evaluation should contain the following elements: Win probability, Threat, and Potential strategies. \\
       & - "final\_evaluation": After all of your thoughts, judge which side takes advantage. 1 means you think 'O' takes advantage, -1 means you think 'X' takes advantage, and 0 means you think the game will be a draw. \\
       & You should output your answer in the JSON format. \\
       & \textbf{EXAMPLE:} \\
       & user: The (board, action) to evaluate is O's turn: \\
       & Board: \\
       & O | O | X \\
       & --------- \\
       & 4 | X | 6 \\
       & --------- \\
       & 7 | 8 | 9 \\
       & Action: The O's move is 4. \\
       & Below is the rollout sequence after this (board, action): \\
       & After O's move 4, the board position is: \\
       & O | O | X \\
       & --------- \\
       & O | X | 6 \\
       & --------- \\
       & 7 | 8 | 9 \\
       & After X's move 7, the board position is: \\
       & O | O | X \\
       & --------- \\
       & O | X | 6 \\
       & --------- \\
       & X | 8 | 9 \\
       & The game is over. X wins. X wins by occupying the positions 3, 5, 7. \\
       & assistant: \{ \\
       & "thought": \{ \\
       & "Reflection": "It appears that the initial board position \\
       & O | O | X \\
       & --------- \\
       & 4 | X | 6 \\
       & --------- \\
       & 7 | 8 | 9 \\
       & and action 4 were not favorable for O, as X was able to block on O's moves and ultimately win the game.", \\
       & "Win probability": "The win probability for X is large, while the win probability for O is low.", \\
       & "Threat": "X has played center 5 and corner 3. X can win by playing corner 7. O was able to occupy 1, 4, 7 and create a potential winning opportunity.", \\
       & "Potential strategies": "Potential strategies for O include playing the corner 7 to block X as opposite corner and win by occupying 1, 4, 7. X could have occupied 3, 5, 7 to win the game. X has already occupied 3, 5, and there is 1 step to complete the plan." \\
       & \}, \\
       & "final\_evaluation": -0.9 \\
       & \} \\ \hline
USER & The board to evaluate is O's turn: \\
     & Board: \\
     & O | X | 3 \\
     & --------- \\
     & 4 | O | 6 \\
     & --------- \\
     & 7 | X | 9 \\
     & Action: The O's move is 9. \\
     & \\
     & Below is the rollout sequence 1 after this (board, action): \\
     & After O taking action 9, the board position is: \\
     & O | X | 3 \\
     & --------- \\
     & 4 | O | 6 \\
     & --------- \\
     & 7 | X | O. \\
     & The game is over. O wins. O wins by occupying the positions [1, 5, 9]. \\
     & \\
     & Below is the rollout sequence 2 after this (board, action): \\
     & After O taking action 9, the board position is: \\
     & O | X | 3 \\
     & --------- \\
     & 4 | O | 6 \\
     & --------- \\
     & 7 | X | O. \\
     & The game is over. O wins. O wins by occupying the positions [1, 5, 9]. \\
     & \\
     & Below is the rollout sequence 3 after this (board, action): \\
     & After O taking action 9, the board position is: \\
     & O | X | 3 \\
     & --------- \\
     & 4 | O | 6 \\
     & --------- \\
     & 7 | X | O. \\
     & The game is over. O wins. O wins by occupying the positions [1, 5, 9]. \\
     & \\
     & Below is the rollout sequence 4 after this (board, action): \\
     & After O taking action 9, the board position is: \\
     & O | X | 3 \\
     & --------- \\
     & 4 | O | 6 \\
     & --------- \\
     & 7 | X | O. \\
     & The game is over. O wins. O wins by occupying the positions [1, 5, 9]. \\
     & \\
     & Below is the rollout sequence 5 after this (board, action): \\
     & After O taking action 9, the board position is: \\
     & O | X | 3 \\
     & --------- \\
     & 4 | O | 6 \\
     & --------- \\
     & 7 | X | O. \\
     & The game is over. O wins. O wins by occupying the positions [1, 5, 9]. \\
     & \\
     & Now generate your evaluation for the (board, action) pair. \\ \hline
ASSISTANT & \{ \\
          & "thought": \{ \\
          & "Reflection": "It appears that the initial board position \\
          & O | X | 3 \\
          & --------- \\
          & 4 | O | 6 \\
          & --------- \\
          & 7 | X | 9 \\
          & and action 9 were highly favorable for O, as O was able to win the game immediately after taking action 9 in all rollout sequences.", \\
          & "Win probability": "The win probability for O is extremely high, while the win probability for X is extremely low.", \\
          & "Threat": "There were no threats from X in any of the rollout sequences, as O was able to win the game immediately after taking action 9.", \\
          & "Potential strategies": "The potential strategy for O was to occupy the corner 9 to win the game by occupying the positions [1, 5, 9]. This strategy was successful in all rollout sequences." \\
          & \}, \\
          & "final\_evaluation": 1 \\
          & \} \\ \hline
\end{longtable}
\end{mdframed}

\textbf{Policy Improvement Details}

Below we show a comparison of the Language Policy Improvement procedure at the beginning of training and at the end. The evaluation is conducted on the following board:

\[
\begin{array}{|c|c|c|}
\hline
\text{O} & \text{X} & 3 \\
\hline
4 & \text{O} & 6 \\
\hline
7 & \text{X} & 9 \\
\hline
\end{array}
\] 

The evolution in policy quality demonstrates how our framework effectively leverages value function assessments to enhance decision-making. The improved policy shows more sophisticated strategic reasoning, considering not just immediate moves but also their implications for future game states. This transformation is achieved while maintaining our restriction against using the model's pre-trained game knowledge, instead relying purely on learned evaluation capabilities.

\begin{mdframed}[style=mystyle, frametitle={Language Policy Improvement (Beginning)}]
\addcontentsline{toc}{paragraph}{Language Policy Improvement (Beginning)}
\small\ttfamily
\begin{longtable}{|p{0.15\textwidth}|p{0.8\textwidth}|}
\hline
\textbf{Role} & \textbf{Content} \\ \hline
SYSTEM & You are playing the game tic-tac-toe on a 3*3 board. Tic Tac Toe is a two-player game played on a grid. \\
       & Players take turns marking a space with their respective symbol. The goal is to get multiple of one’s own symbols in a row, either horizontally, vertically, or diagonally, before the opponent does. If all nine squares are filled and no player has three in a row, the game is a draw. \\
       & The board consists of "O", "X" and grid number. The grid number indicates an empty grid. \\
       & Your task is to determine the best move for the next player based on the given board position and the next player. \\
       & The evaluations of (board, action) pairs after possible moves are given. \\
       & \textbf{DO NOT} judge the board based on your knowledge, only use the evaluations to determine the best move. \\
       & The evaluation for the next board is in the format of a JSON, consisting of two elements: \\
       & - \textbf{"thought"}: Evaluation of the board and action pair. \\
       & - \textbf{"final\_evaluation"}: Judge which side takes advantage. 1 means 'O' takes advantage, -1 means 'X' takes advantage, and 0 means the game will be a draw. \\ \hline
USER & Here is the board position and the next player is O: \\
     & Board: \\
     & O | X | 3 \\
     & --------- \\
     & 4 | O | 6 \\
     & --------- \\
     & 7 | X | 9. \\
     & The possible moves are [3, 7, 4, 9]. \\
     & The following are the boards after each possible move: \\
     & \textbf{\#\#\# Evaluation for taking action 3:} \\
     & \{ \\
     & "thought": \{ \\
     & "Reflection": "The initial board position \\
     & O | X | 3 \\
     & --------- \\
     & 4 | O | 6 \\
     & --------- \\
     & 7 | X | 9 \\
     & and action 3 were favorable for O, as O was able to win the game in all rollout sequences.", \\
     & "Win probability": "The win probability for O is high, while the win probability for X is low.", \\
     & "Threat": "O has played corner 1 and center 5. O can win by playing corner 3, 4, 7 or 9. X has played corner 2 and 8 but failed to block O's winning lines.", \\
     & "Potential strategies": "Potential strategies for O include playing the corner 3 to create a potential winning opportunity. O could have occupied 1, 4, 7 or 1, 5, 9 or 3, 5, 7 to win the game. X could have blocked O's winning lines by playing 4 or 7 but failed to do so." \\
     & \}, \\
     & "final\_evaluation": 0.9 \\
     & \} \\
     & \textbf{\#\#\# Evaluation for taking action 7:} \\
     & \{ \\
     & "thought": \{ \\
     & "Reflection": "The initial board position \\
     & O | X | 3 \\
     & --------- \\
     & 4 | O | 6 \\
     & --------- \\
     & 7 | X | 9 \\
     & and action 7 were favorable for O, as O was able to win the game in all rollout sequences.", \\
     & "Win probability": "The win probability for O is high, while the win probability for X is low.", \\
     & "Threat": "O has played corner 1 and center 5. O can win by playing corner 7 and then 9, or 4 and then 7. X has played corner 3 and 8 but failed to block O's winning lines.", \\
     & "Potential strategies": "Potential strategies for O include playing the corner 7 to create a potential winning opportunity and then playing 9 or 4 to win the game. X could have occupied 3, 6, 9 to block O's winning lines but failed to do so." \\
     & \}, \\
     & "final\_evaluation": 0.9 \\
     & \} \\
     & \textbf{\#\#\# Evaluation for taking action 4:} \\
     & \{ \\
     & "thought": \{ \\
     & "Reflection": "The initial board position \\
     & O | X | 3 \\
     & --------- \\
     & 4 | O | 6 \\
     & --------- \\
     & 7 | X | 9 \\
     & and action 4 were favorable for O, as O was able to win the game in all rollout sequences.", \\
     & "Win probability": "The win probability for O is high, while the win probability for X is low.", \\
     & "Threat": "O has played center 5 and corner 1. O can win by playing corner 7 or 9. X has played center 6 and corner 3 but failed to block O's winning lines.", \\
     & "Potential strategies": "Potential strategies for O include playing the corner 7 or 9 to win the game. X could have occupied 3, 6, 9 to block O's winning lines but failed to do so." \\
     & \}, \\
     & "final\_evaluation": 0.9 \\
     & \} \\
     & \textbf{\#\#\# Evaluation for taking action 9:} \\
     & \{ \\
     & "thought": \{ \\
     & "Reflection": "It appears that the initial board position \\
     & O | X | 3 \\
     & --------- \\
     & 4 | O | 6 \\
     & --------- \\
     & 7 | X | 9 \\
     & and action 9 were favorable for O, as O was able to win the game in both rollout sequences.", \\
     & "Win probability": "The win probability for O is high, while the win probability for X is low.", \\
     & "Threat": "O has played center 5 and corner 1. O can win by playing corner 9. X has played corner 3 and center 6 is empty, but X was not able to block O's moves and ultimately lost the game.", \\
     & "Potential strategies": "Potential strategies for O include playing the corner 9 to create a potential winning opportunity and then playing 3 or 7 to win the game. X could have occupied 3, 6, 7 to block O's winning opportunity, but X was not able to do so." \\
     & \}, \\
     & "final\_evaluation": 0.9 \\
     & \} \\
     & Now, please give your evaluation and the best move for O based on the given board position: \\
     & O | X | 3 \\
     & --------- \\
     & 4 | O | 6 \\
     & --------- \\
     & 7 | X | 9. \\
     & You should output your answer in the JSON format. Your answer consists of two elements: \\
     & - \textbf{"thought"}: Let's think step by step. Generate your detailed reflection by analyzing the next board positions and their evaluations. \\
     & - \textbf{"best\_move"}: The best move for the next player. The move should be in the format of a number from 1 to 9, indicating the position on the board. And the move should be in the available positions [3, 7, 4, 9]. \\
     & Don't output extra information except for the JSON format. \\ \hline
ASSISTANT & \{ \\
          & "thought": "Analyzing the given board position \\
          & O | X | 3 \\
          & --------- \\
          & 4 | O | 6 \\
          & --------- \\
          & 7 | X | 9 and the evaluations of the possible moves, we can see that all four moves (3, 7, 4, 9) have the same final evaluation of 0.9, indicating that O has a high chance of winning with any of these moves. However, we should consider the potential strategies and threats for each move. \\
          & Move 3 and 7 allow O to create a potential winning opportunity by playing the corner, while move 4 allows O to play the center and create a potential winning opportunity by playing the corner 7 or 9. Move 9 allows O to play the corner and create a potential winning opportunity by playing 3 or 7. \\
          & Considering the potential strategies, all four moves seem to have similar advantages. However, move 7 seems to have a slight advantage as it allows O to create a potential winning opportunity and then play 9 to win the game, while also putting pressure on X to block the winning line. \\
          & Therefore, the best move for O is 7.", \\
          & "best\_move": 7 \\
          & \} \\ \hline
\end{longtable}
\end{mdframed}

\begin{mdframed}[style=mystyle, frametitle={Language Policy Improvement (End)}]
\addcontentsline{toc}{paragraph}{Language Policy Improvement (End)}
\small\ttfamily
\begin{longtable}{|p{0.15\textwidth}|p{0.8\textwidth}|}
\hline
\textbf{Role} & \textbf{Content} \\ \hline
SYSTEM & You are playing the game tic-tac-toe on a 3*3 board. Tic Tac Toe is a two-player game played on a grid. \\
       & Players take turns marking a space with their respective symbol. The goal is to get multiple of one’s own symbols in a row, either horizontally, vertically, or diagonally, before the opponent does. If all nine squares are filled and no player has three in a row, the game is a draw. \\
       & The board consists of "O", "X" and grid numbers. The grid number indicates an empty grid. \\
       & Your task is to determine the best move for the next player based on the given board position and the next player. \\
       & The evaluations of (board, action) pairs after possible moves are given. \\
       & \textbf{DO NOT} judge the board based on your knowledge, only use the evaluations to determine the best move. \\
       & The evaluation for the next board is in the format of a JSON, consisting of two elements: \\
       & - \textbf{"thought"}: Evaluation of the board and action pair. \\
       & - \textbf{"final\_evaluation"}: Judge which side takes advantage. 1 means 'O' takes advantage, -1 means 'X' takes advantage, and 0 means the game will be a draw. \\ \hline
USER & Here is the board position and the next player is O: \\
     & Board: \\
     & O | X | 3 \\
     & --------- \\
     & 4 | O | 6 \\
     & --------- \\
     & 7 | X | 9. \\
     & The possible moves are [9]. \\
     & The following are the boards after each possible move: \\
     & \textbf{\#\#\# Evaluation for taking action 9:} \\
     & \{ \\
     & "thought": \{ \\
     & "Reflection": "It appears that the initial board position \\
     & O | X | 3 \\
     & --------- \\
     & 4 | O | 6 \\
     & --------- \\
     & 7 | X | 9 \\
     & and action 9 were highly favorable for O, as O was able to win the game immediately after taking action 9 in all rollout sequences.", \\
     & "Win probability": "The win probability for O is extremely high, while the win probability for X is extremely low.", \\
     & "Threat": "There were no threats from X in any of the rollout sequences, as O was able to win the game immediately after taking action 9.", \\
     & "Potential strategies": "The potential strategy for O was to occupy the corner 9 to win the game by occupying the positions [1, 5, 9]. This strategy was successful in all rollout sequences." \\
     & \}, \\
     & "final\_evaluation": 1 \\
     & \} \\
     & Now, please give your evaluation and the best move for O based on the given board position: \\
     & O | X | 3 \\
     & --------- \\
     & 4 | O | 6 \\
     & --------- \\
     & 7 | X | 9. \\
     & You should output your answer in the JSON format. Your answer consists of two elements: \\
     & - \textbf{"thought"}: Let's think step by step. Generate your detailed reflection by analyzing the next board positions and their evaluations. \\
     & - \textbf{"best\_move"}: The best move for the next player. The move should be in the format of a number from 1 to 9, indicating the position on the board. And the move should be in the available positions [9]. \\
     & Don't output extra information except for the JSON format. \\ \hline
ASSISTANT & \{ \\
          & "thought": "Based on the evaluation of the board position after taking action 9, it appears that this move is highly favorable for O. The win probability for O is extremely high, while the win probability for X is extremely low. There were no threats from X in any of the rollout sequences, as O was able to win the game immediately after taking action 9. The potential strategy for O was to occupy the corner 9 to win the game by occupying the positions [1, 5, 9]. This strategy was successful in all rollout sequences. Therefore, taking action 9 is the best move for O.", \\
          & "best\_move": 9 \\
          & \} \\ \hline
\end{longtable}
\end{mdframed}

\subsubsection{Baseline Implementation Details}
For our PPO baseline, we implement a custom PPO agent adapted for language models. The agent consists of two key components: an actor network based on LLaMA-3.1-8B-Instruct for action selection, and a critic network for value estimation.

\paragraph{Architecture Details}
The actor network uses \textit{LLaMA-3.1-8B-Instruct}'s architecture with added special tokens for moves 1-9. The critic network consists of a two-layer MLP (Multi-Layer Perceptron) that takes the last hidden state from the actor as input:
\begin{itemize}
    \item Input dimension: \textit{LLaMA-3.1-8B} hidden size
    \item Hidden layer: ReLU activation
    \item Output layer: Single value with Tanh activation
\end{itemize}

\paragraph{Training Configuration}
The PPO training uses the following hyperparameters:
\begin{table}[htbp]
\centering
\begin{tabular}{@{}ll@{}}
\toprule
\textbf{Parameter} & \textbf{Value} \\
\midrule
Learning Rate (Actor) & 1e-5 \\
Learning Rate (Critic) & 1e-5 \\
PPO Epochs & 1 \\
Batch Size & 16 \\
Gradient Accumulation Steps & 2 \\
Value Loss Coefficient & 0.5 \\
Entropy Coefficient & 0.01 \\
PPO Clip Range & 0.2 \\
Discount Factor ($\gamma$) & 0.99 \\
GAE Parameter ($\lambda$) & 0.95 \\
\bottomrule
\end{tabular}
\caption{PPO Training Hyperparameters}
\label{tab:ppo_hyperparameters}
\end{table}

\paragraph{Implementation Details}
For each training iteration:
\begin{itemize}
    \item We collect 512 complete game trajectories using 8 parallel environments
    \item Actions are selected using a temperature-controlled sampling ($T=0.7$) over the logits
    \item The policy proposes $N_{sample}=1$ candidate actions per state and selects top-$k$ ($k=1$) based on sampling frequency
    \item We use 1 A100 GPU for training
    \item Gradient checkpointing is enabled to optimize memory usage
\end{itemize}

Both networks are trained using AdamW optimizer with weight decay of 0.01. We maintain experience buffers with a history length of 1 iterations for on-policy optimization.

\subsubsection{Additional Experiments}\label{app:tictactoe:first_move}

\begin{figure}[H]
    \vspace{-0pt}
    \centering
    \includegraphics[width=0.9\linewidth]{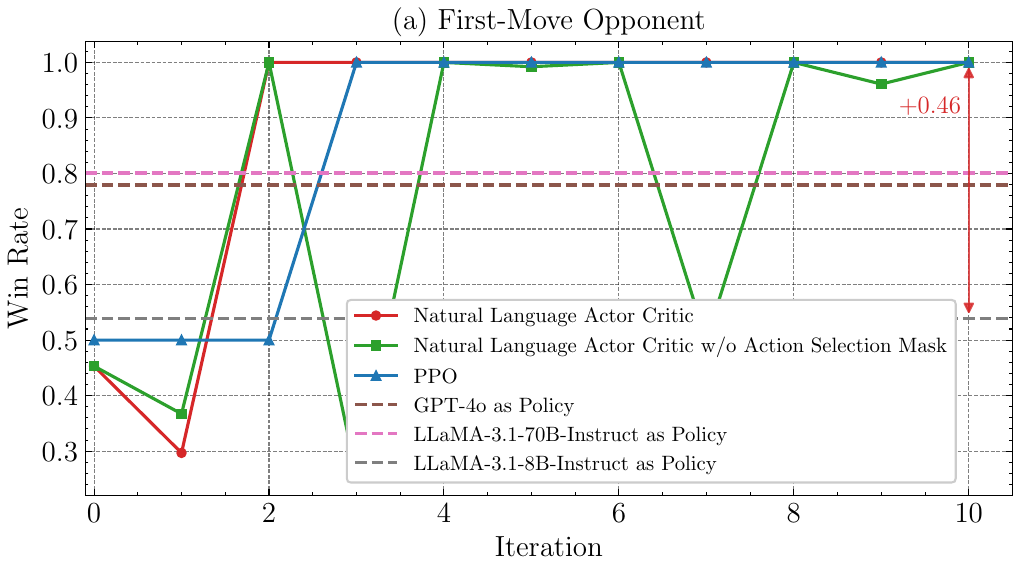}
    \vspace{-5pt}
    \caption{Training results against the deterministic first-move opponent.}
    \vspace{-13pt}
    \label{fig:ab_first_move}
\end{figure}

\subsubsection{Additional Ablation Studies}\label{app:tictactoe:topm}

\textbf{Action Selection Parameter Study}
To complement the main ablation studies presented in Section~\ref{exp:nlrl_hparam_abl}, we conduct additional experiments on the action selection parameter top-$m$. Our results demonstrate that increasing $m$ from 2 to 10 enhances exploration capabilities while maintaining training stability. Specifically, larger $m$ value (10) show more stable training compared to smaller values (2-5).

\begin{figure}[H]
    \vspace{-0pt}
    \centering
    \includegraphics[width=0.9\linewidth]{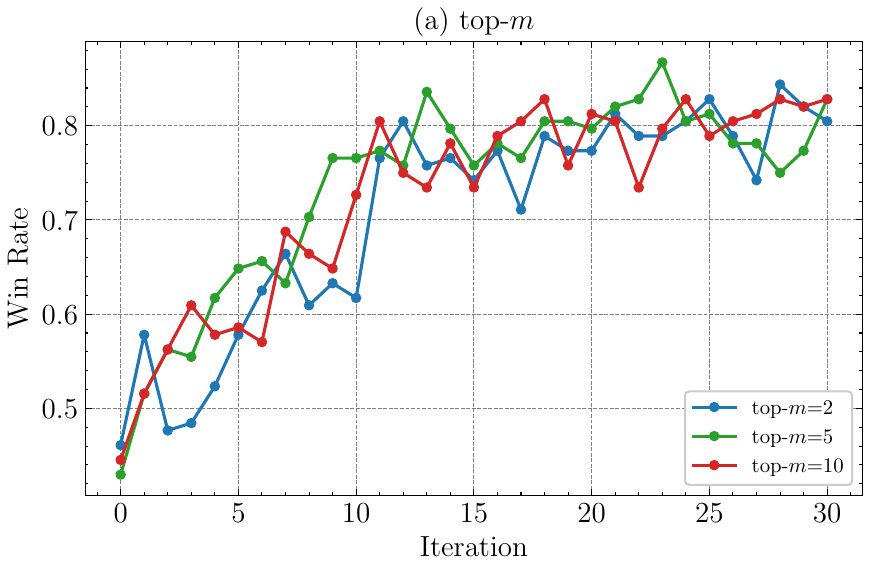}
    \vspace{-5pt}
    \caption{Ablation studies on action selection parameter  top-$m$.}
    \vspace{-13pt}
    \label{fig:ab_topm_exp}
\end{figure}

\subsection{FrozenLake Experiment}
\label{app:frozenlake}

FrozenLake environment is a grid-based navigation task (e.g., 4×4 grids) where a player must reach a goal while avoiding holes. The board looks like this:
\begin{figure}[h]
    \centering
    \begin{minipage}{0.6\textwidth}
        \centering
        \[
        \begin{array}{|cccc|}
        \hline
        \text{\_} & \text{G} & \text{O} & \text{\_} \\
        \text{\_} & \text{\_} & \text{\_} & \text{O} \\
        \text{P} & \text{O} & \text{\_} & \text{\_} \\
        \text{\_}& \text{\_} & \text{\_} & \text{\_}\\
        \hline
        \end{array}
        \]
    \end{minipage}
    \begin{minipage}{0.35\textwidth}
        \centering
        \includegraphics[width=\linewidth]{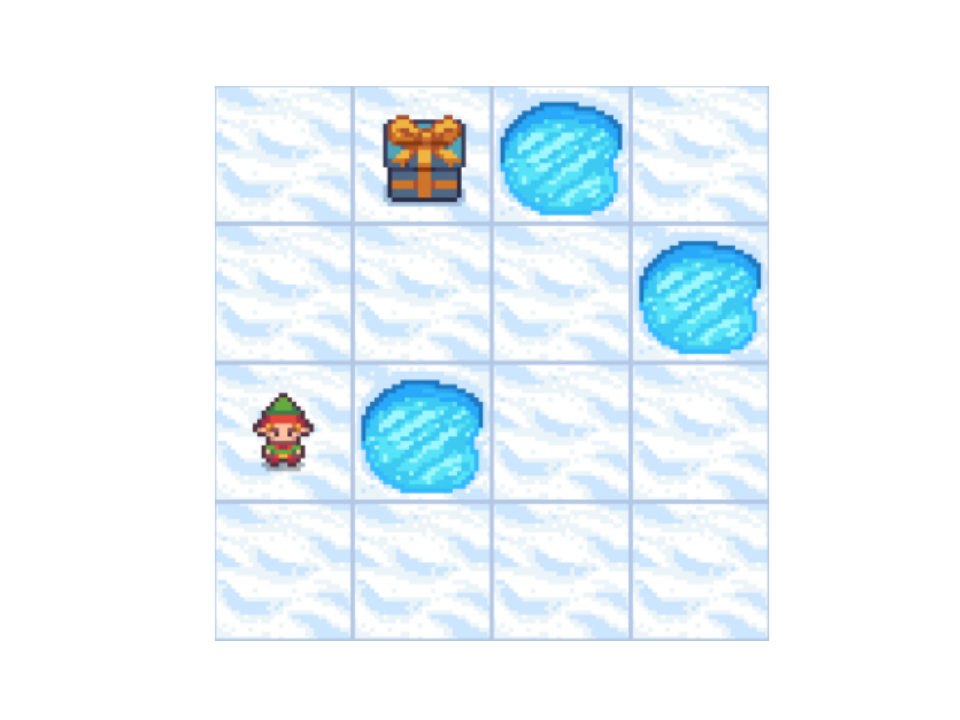}
    \end{minipage}
\end{figure}

Since it has similar properties to the tic-tac-toe game, we borrow most of the implementation and hyperparameter settings introduced in Appendix \ref{app:tictactoe}. In FrozenLake experiments, we specifically compare NLRL with RAGEN \cite{wang2025ragen}. The implementation of RAGEN follows its original paper and open codebase: \url{https://github.com/RAGEN-AI/RAGEN} (MIT License) with minimal modification. The original RAGEN paper has only tried learning on Qwen2.5-0.5B-Instruction model, whereas we also extend it to other backbone language models to investigate the CoT degradation phenomenon.

\subsubsection{Comparing Detailed Reasoning}
We have tried RAGEN-like training on multiple model series including Qwen2.5-0.5B, Qwen2.5-7B and Llama3.1-8B. All of them have experienced a degradation in reasoning, as illustrated in Figure \ref{tab:ragen_vs_nlrl_full}.
\begin{table}[htbp]
  \centering
  \renewcommand{\arraystretch}{1.5}
  \begin{tabular}{|m{1.5cm}|m{0.8cm}|m{4cm}|m{6cm}|}
    \hline
    \textbf{Methods} & \textbf{Map} & \textbf{Initial Stage} & \textbf{Late Training} \\
    \hline
    {RAGEN} (\textit{Qwen2.5-0.5B-Instruct}) & 
    \multirow{3}{*}{\raisebox{-3.5cm}{\begin{tabular}{@{}l@{}}
      \texttt{\_G0\_} \\
      \texttt{\_\_\_O} \\
      \texttt{PO\_\_} \\
      \texttt{\_\_\_\_}
    \end{tabular}}} & 
    \small \texttt{(THINK):} Let's start by considering the initial state: \textit{<MAP>}. The goal is to reach the target 'PO\_\_\_' from the hole 'O\_\_O\_\_O\_\_O'. I'll start by moving to the left to avoid the hole. \newline 
    \small \texttt{(Action):} Left || Up & 
    \small \texttt{(THINK):} Forbid the goal and go left then up. \newline 
    \small \texttt{(Action):} Left || Up \\
    \cline{1-1} \cline{3-4}
    {RAGEN} (\textit{Qwen2.5-7B-Instruct}) &  & 
    \small \texttt{(THINK):} Move left to avoid the hole and then go up to reach the goal. \newline 
    \small \texttt{(Action):} Left || Up & 
    \small \texttt{(THINK):} Up may lead to the hole or the target. \newline 
    \small \texttt{(Action):} Up || Up \\
    \cline{1-1} \cline{3-4}
    {RAGEN} (\textit{Llama3.1-8B-Instruct}) &
    & 
    \small \texttt{(THINK):} To avoid the hole and go to the target, I should go up to stay safe and then move to the right to approach the goal from a safe distance. \newline 
    \small \texttt{(Action):} Up || Right & 
    \small \texttt{(THINK):} My thoughts are to go to the goal \newline 
    \small \texttt{(Action):} Up || Right \\
    \cline{1-1} \cline{3-4}
    \textbf{NLRL} (\textit{Llama3.1-8B-Instruct}) & & 
    \small \texttt{(THINK):} The current board position is <\textit{MAP}>. The goal is on the top left, and the player is on the bottom left. The player is not in the hole. The available moves are 1, 2, 3, 4. Since the player is already on the leftmost position, moving left is not an option. The player can move down to get closer to the goal. Therefore, the best next move for P is to play action 2 to move down.  \newline 
    \small \texttt{(Action):} Down & 
    \small \texttt{(THINK):} Based on the evaluations of the next board positions, moving up results in the highest final evaluation of +5, indicating that P is one block closer to the goal position G after the move. Although there is a hole one block away from P after moving up, the potential strategy of moving up and then moving right can help P avoid the hole and arrive at the goal position. In contrast, moving left does not immediately change P's position, and moving down  moves P away from the goal position G. Therefore, moving up is the most favourable move. \newline 
    \small \texttt{(Action):} Up \\
    \hline
  \end{tabular}
  \vspace{0.3cm}
  \caption{Comparing detailed reasoning on FrozenLake environment between RAGEN \citep{wang2025ragen} and NLRL (Ours). No matter what backbone model we used, RAGEN style of training will downgrade the model only to output the simplest reasoning. Our method encourages the model to develop rigorous reasoning as the training progresses.}
  \label{tab:ragen_vs_nlrl_full}
\end{table}

\subsubsection{Visualization}

\begin{mdframed}[style=mystyle, frametitle={Language State-action Value (Before)}]
\addcontentsline{toc}{paragraph}{Language State-action Value (Before)}
\small\ttfamily
\begin{longtable}{|p{0.15\textwidth}|p{0.8\textwidth}|}
\hline
\textbf{Role} & \textbf{Content} \\ \hline
SYSTEM & You are an expert agent at playing the game FrozenLake on a 4*4 board. FrozenLake is a single-player game played on a grid. The goal is to avoid the hole and move to the goal. You may move to the unintended direction due to the slippery ice. \\
& \\
& The meaning of each symbol in the state is: \\
& P: player, \_: empty, O: hole, G: goal, X: player in hole, $\surd$ : player on goal \\
& \\
& The possible actions are: \\
&   1:Left, 2:Down, 3:Right, 4:Up \\
& \\
& You are learning how to evaluate a (board, action) pair in the FrozenLake by playing the game given the (board, action) pair and reflect the playing history. \\
& The playing history depicts a game in progress with a final result. Your answer consists of two elements: \\
& - "thought": let's think step by step. Generate your detailed evaluation over the (board, action) pair by merely reflecting the playing history after this pair from different perspectives. You should only rely on the playing history as context and don't evaluate the game with your own judgment. Your evaluation should contain the following elements: Win probability, Threat, Potential strategies. \\
& "final\_evaluation": After all your thought, finally judge how good the agent's current position is compared to all the possible positions in the board, in terms of reaching the goal. Measure the goodness in terms of negative distance to the goal position after taking this action. For example, 4 blocks away will gives -0.4; 1 blocks away will gives -0.1; if the agent fall into the hole, gives -5; if the agent has reach the goal, gives 5. \\
& \\
& \#\#\# EXAMPLE:\\
& user: The (board, action) to evaluate: \\
&Board: \\
&P\_\_\_ \\
&\_O\_\_ \\
&\_\_\_O \\
&GO\_\_ \\
&Action: \\
&2 (Down). \\
& \\
& For your reference, below is the rollout sequence 1 after this (board, action): \\
& After move 2 (Down), the board position is: \\
&\_\_\_\_ \\
&PO\_\_ \\
&\_\_\_O \\
&GO\_\_ \\
& \\
&After move 3 (Right), the board position is: \\
&\_\_\_\_ \\
&\_X\_\_ \\
&\_\_\_O \\
&GO\_\_ \\
&The game is over. Player fall into the hole and therefore fails. \\
& \\
& assistant: \\
& \{"thought": \{"Reflection": "It appears that the initial board position \\
& P\_\_\_ \\
&\_O\_\_ \\
&\_\_\_O \\
&GO\_\_ \\
&and move 2 (Down) \\
&was favorable for P, as P is one block closer to the goal position G.", "Win probability": "The win/success probability for P is larger.", "Threat": "There are three holes on the board. There is a hole only one block away from P after action has been taken, which can be dangerous if O choose to move right.", "Potential strategies": "Potential strategies for P include moving 2 (Down) twice to arrive at goal position."\} \\
& "final\_evaluation": -0.3\}\\
 \hline
USER & The board to evaluate is: \\
&Board: \\
&\_OO\_ \\
&G\_\_\_ \\
&\_\_\_\_ \\
&\_OOP \\
&Action: \\
&The next move is 2. \\
& \\
&For your reference, below is the rollout sequence 1 after this (board, action): \\
&After taking action 2, the board position is \\
&\_OO\_\\
&G\_\_\_\\
&\_\_\_\_\\
&\_OOP.\\
&\\
&After taking action 2, the board position is \\
&\_OO\_ \\
&G\_\_\_ \\
&\_\_\_\_ \\
&\_OOP. \\
& \\
&...... \\
& \\
&The game is over. Player fall into the hole and therefore fails. \\
& \\
&Now generate your evaluation for the (board, action) pair.\\
\hline
ASSISTANT & \{"thought": \{"Reflection": "It appears that the initial board position \\
&\_OO\_ \\
&G\_\_\_ \\
&\_\_\_\_ \\
&\_OOP \\
&and move 2 (Down) \\ 
&was not favorable for P, as P did not move after taking this action in both rollout sequences.", "Win probability": "The win/success probability for P is smaller.", "Threat": "There are four holes on the board. There are holes both above and below P after action has been taken, which can be very dangerous if P choose to move up or down.", "Potential strategies": "Potential strategies for P include moving 3 (Right) or 1 (Left) to avoid the holes above and below P."\} \\
& "final\_evaluation": -0.5\} \\
\hline
\end{longtable}
\end{mdframed}

\begin{mdframed}[style=mystyle, frametitle={Language State-action Value (After)}]
\addcontentsline{toc}{paragraph}{Language State-action Value (Before)}
\small\ttfamily
\begin{longtable}{|p{0.15\textwidth}|p{0.8\textwidth}|}
\hline
\textbf{Role} & \textbf{Content} \\ \hline
SYSTEM & You are an expert agent at playing the game FrozenLake on a 4*4 board. FrozenLake is a single-player game played on a grid. The goal is to avoid the hole and move to the goal. You may move to the unintended direction due to the slippery ice. \\
        & \\
        & The meaning of each symbol in the state is: \\
        & P: player, \_: empty, O: hole, G: goal, X: player in hole, $\surd$ : player on goal \\
        & \\
        & The possible actions are: \\
        &   1:Left, 2:Down, 3:Right, 4:Up \\
        & \\
        & You are learning how to evaluate a (board, action) pair in the FrozenLake by playing the game given the (board, action) pair and reflect the playing history. \\
        & The playing history depicts a game in progress with a final result. Your answer consists of two elements: \\
        & - "thought": let's think step by step. Generate your detailed evaluation over the (board, action) pair by merely reflecting the playing history after this pair from different perspectives. You should only rely on the playing history as context and don't evaluate the game with your own judgment. Your evaluation should contain the following elements: Win probability, Threat, Potential strategies. \\
        & "final\_evaluation": After all your thought, finally judge how good the agent's current position is compared to all the possible positions in the board, in terms of reaching the goal. Measure the goodness in terms of negative distance to the goal position after taking this action. For example, 4 blocks away will gives -0.4; 1 blocks away will gives -0.1; if the agent fall into the hole, gives -5; if the agent has reach the goal, gives 5. \\
        & \\
        & \#\#\# EXAMPLE:\\
        & user: The (board, action) to evaluate: \\
        &Board: \\
        &P\_\_\_ \\
        &\_O\_\_ \\
        &\_\_\_O \\
        &GO\_\_ \\
        &Action: \\
        &2 (Down). \\
        & \\
        & For your reference, below is the rollout sequence 1 after this (board, action): \\
        & After move 2 (Down), the board position is: \\
        &\_\_\_\_ \\
        &PO\_\_ \\
        &\_\_\_O \\
        &GO\_\_ \\
        & \\
        &After move 3 (Right), the board position is: \\
        &\_\_\_\_ \\
        &\_X\_\_ \\
        &\_\_\_O \\
        &GO\_\_ \\
        &The game is over. Player fall into the hole and therefore fails. \\
        & \\
        & assistant: \\
        & \{"thought": \{"Reflection": "It appears that the initial board position \\
        & P\_\_\_ \\
        &\_O\_\_ \\
        &\_\_\_O \\
        &GO\_\_ \\
        &and move 2 (Down) \\
        &was favorable for P, as P is one block closer to the goal position G.", "Win probability": "The win/success probability for P is larger.", "Threat": "There are three holes on the board. There is a hole only one block away from P after action has been taken, which can be dangerous if O choose to move right.", "Potential strategies": "Potential strategies for P include moving 2 (Down) twice to arrive at goal position."\} \\
        & "final\_evaluation": -0.3\}\\
 \hline
USER & The board to evaluate is: \\
        Board: \\
        &\_\_\_P \\
        &\_OO\_ \\
        &\_O\_\_ \\
        &\_\_G\_ \\
        &Action: \\
        &The next move is 3. \\
        & \\
        &For your reference, below is the rollout sequence 1 after this (board, action): \\
        &After taking action 3, the board position is \\
        &\_\_\_P \\
        &\_OO\_ \\
        &\_O\_\_ \\
        &\_\_G\_. \\
        & \\
        &After taking action 3, the board position is \\
        &\_\_\_P \\ 
        &\_OO\_ \\
        &\_O\_\_ \\
        &\_\_G\_. \\
        &The game is over. Player has reach maximum number of move and therefore fails. \\
        &\\
        &......\\
        & \\
        &Now generate your evaluation for the (board, action) pair. \\
\hline
ASSISTANT & \{"thought": \{"Reflection": "It appears that the initial board position \\
        &\_\_\_P \\
        &\_OO\_ \\
        &\_O\_\_ \\
        &\_\_G\_ \\
        &and move 3 (Right) \\ 
        &resulted in no change in the board position in all rollout sequences.", "Win probability": "The win/success probability for P is low, as P did not move closer to the goal position G in any rollout sequence.", "Threat": "There are four holes on the board. However, none of the rollout sequences resulted in P falling into a hole.", "Potential strategies": "Potential strategies for P include moving down to avoid the holes and get closer to the goal position."\} \\
        &"final\_evaluation": -0.4\} \\
 \hline
\end{longtable}
\end{mdframed}

\begin{mdframed}[style=mystyle, frametitle={Language Policy Improvement (Beginning)}]
\addcontentsline{toc}{paragraph}{Language Policy Improvement (Beginning)}
\small\ttfamily
\begin{longtable}{|p{0.15\textwidth}|p{0.8\textwidth}|}
\hline
\textbf{Role} & \textbf{Content} \\ \hline
SYSTEM & You are an expert agent at playing the game FrozenLake on a 4*4 board. FrozenLake is a single-player game      played on a grid.
        The goal is to avoid the hole and move to the goal. You may move to the unintended direction due to the slippery ice.\\
        & \\
        & Your task is to evaluate a given board position.\\
        &The meaning of each symbol in the state is:\\
        &P: player, \_: empty, O: hole, G: goal, X: player in hole, $\surd$: player on goal \\
        &\\
        &The possible actions are:\\
        &1:Left, 2:Down, 3:Right, 4:Up\\
        &\\
        &Your task is to determine the best next move based on the given board position.\\
        &The evaluations of (board, action) pairs after possible moves are given.\\
        &DO NOT judge the board based on your knowledge, only use the evaluations to determine the best move.\\
        &The evaluation for the next board is in the format of a json format, consisting of two elements:\\
        &- "thought": evaluation of the the board and action pair.\\
        &- "final\_evaluation": After all your thought, finally judge how good the agent's position is compared to all the possible positions in the board, in terms of reaching the goal. Measure the goodness in terms of negative distance to the goal position. For example, 4 blocks away will gives -0.4; 1 blocks away will gives -0.1; if the agent fall into the hole, gives -5; if the agent has reach the goal, gives 5.\\
\hline
USER & Here is the board position:\\
        &\_\_P\_ \\
        &\_G\_\_ \\
        &\_\_\_\_ \\
        &\_\_\_\_. The possible moves are [2, 3, 4]. \\
        &The following are the boards after each possible move: \\
        &\\
        &\#\#\# Evaluation for taking action 2: \\
        &\{"thought": \{"Reflection": "It appears that the initial board position \\
        &\_\_P\_ \\
        &\_G\_\_ \\
        &\_\_\_\_ \\
        &\_\_\_\_ \\
        &and move 2 (Down) \\ 
        &was favorable for P, as P is one block closer to the goal position G after taking this action.", "Win probability": "The win/success probability for P is larger.", "Threat": "There are no holes on the board that are immediately threatening P after the action has been taken.", "Potential strategies": "Potential strategies for P include moving 2 (Down) to arrive at goal position, and then moving left to reach the goal."\} \\
        &"final\_evaluation": 0.1\} \\
        &\\
        &\#\#\# Evaluation for taking action 3: \\
        &\{"thought": \{"Reflection": "It appears that the initial board position \\
        &\_\_P\_ \\
        &\_G\_\_ \\
        &\_\_\_\_ \\
        &\_\_\_\_ \\
        &and move 3 (Right)  \\
        &was not favorable for P, as P is now further away from the goal position G.", "Win probability": "The win/success probability for P is smaller.", "Threat": "There are no holes on the board that are immediately threatening P after the action has been taken.", "Potential strategies": "Potential strategies for P could have been to move 2 (Down) to arrive at goal position, but the chosen action 3 (Right) did not lead to this outcome."\} \\
        &"final\_evaluation": -0.4\} \\
        & \\
        &\#\#\# Evaluation for taking action 4: \\
        &\{"thought": \{"Reflection": "It appears that the initial board position \\
        &\_\_P\_ \\
        &\_G\_\_ \\ 
        &\_\_\_\_ \\
        &\_\_\_\_ \\
        &and move 4 (Up) \\
        &resulted in no change in the board position, which indicates that the move was not effective in making progress towards the goal.", "Win probability": "The win/success probability for P is lower.", "Threat": "There are no holes on the board that are immediately threatening P after the action has been taken.", "Potential strategies": "Potential strategies for P include moving 2 (Down) to move closer to the goal position G, but the rollout sequence shows that P did not take this opportunity and instead moved in other directions, which did not lead to a successful outcome."\}\\
        &"final\_evaluation": -0.4\}\\
        &\\
        &Now, please give your evaluation and the best next move based on the given board position \\
        &\_\_P\_\\
        &\_G\_\_\\
        &\_\_\_\_\\
        &\_\_\_\_.\\
        &You should output your answer in the json format. Your answer consists of two elements:\\
        &- "thought": let's think step by step. Generate your detailed reflection by analyzing the next board positions and 
         their evaluations.\\
        &- "best\_move": the best move for the next player. The move should be in the format of a number from 1 to 4, indicating the moving direction.\\
        &Don't output extra information except for the json format.\\ 
 \hline
ASSISTANT & \{ \\
          &"thought" : "Analyzing the given board position \_\_P\_ \_G\_\_ \_\_\_\_ \_\_\_\_ and the evaluations of the possible moves, it appears that taking action 2 (Down) is the most favorable move. This is because the evaluation for this move indicates that P is one block closer to the goal position G and the win/success probability for P is larger. In contrast, taking action 3 (Right) or action 4 (Up) would not be effective in making progress towards the goal, as P would be further away from the goal position G or the board position would remain unchanged. Therefore, the best next move is to take action 2 (Down) to move closer to the goal position G.", \\
          &"best\_move" : 2 \\
        &\}\\
\hline
\end{longtable}
\end{mdframed}

\begin{mdframed}[style=mystyle, frametitle={Language Policy Improvement (End)}]
\addcontentsline{toc}{paragraph}{Language Policy Improvement (Beginning)}
\small\ttfamily
\begin{longtable}{|p{0.15\textwidth}|p{0.8\textwidth}|}
\hline
\textbf{Role} & \textbf{Content} \\ \hline
SYSTEM & You are an expert agent at playing the game FrozenLake on a 4*4 board. FrozenLake is a single-player game      played on a grid.
        The goal is to avoid the hole and move to the goal. You may move to the unintended direction due to the slippery ice.\\
        & \\
        & Your task is to evaluate a given board position.\\
        &The meaning of each symbol in the state is:\\
        &P: player, \_: empty, O: hole, G: goal, X: player in hole, $\surd$: player on goal \\
        &\\
        &The possible actions are:\\
        &1:Left, 2:Down, 3:Right, 4:Up\\
        &\\
        &Your task is to determine the best next move based on the given board position.\\
        &The evaluations of (board, action) pairs after possible moves are given.\\
        &DO NOT judge the board based on your knowledge, only use the evaluations to determine the best move.\\
        &The evaluation for the next board is in the format of a json format, consisting of two elements:\\
        &- "thought": evaluation of the the board and action pair.\\
        &- "final\_evaluation": After all your thought, finally judge how good the agent's position is compared to all the possible positions in the board, in terms of reaching the goal. Measure the goodness in terms of negative distance to the goal position. For example, 4 blocks away will gives -0.4; 1 blocks away will gives -0.1; if the agent fall into the hole, gives -5; if the agent has reach the goal, gives 5.\\
\hline
USER & Here is the board position:\\
        &\_\_G\_ \\
        &\_\_\_O \\
        &\_O\_\_ \\
        &P\_O\_. The possible moves are [4, 1, 3]. \\
        &The following are the boards after each possible move: \\
        &\\
        &\#\#\# Evaluation for taking action 4: \\
        &\{"thought": \{"Reflection": "It appears that the initial board position  \\
        &\_\_G\_ \\
        &\_\_\_O \\
        &\_O\_\_ \\
        &P\_O\_ \\
        &and move 4 (Up) \\
        &was not favorable for P, as P is still two blocks away from the goal position G and there are holes around P.", "Win probability": "The win/success probability for P is smaller.", "Threat": "There are three holes on the board. There are two holes only one block away from P after action has been taken, which can be very dangerous if P choose to move right or left.", "Potential strategies": "Potential strategies for P include moving 4 (Up) again to arrive at the middle row and then move 3 (Right) twice to arrive at goal position, but the rollout sequence shows that this strategy is not successful."\} \\
        &"final\_evaluation": -0.5\} \\
        & \\
        &\#\#\# Evaluation for taking action 1: \\
        &\{"thought": \{"Reflection": "It appears that the initial board position \\
        &\_\_G\_ \\
        &\_\_\_O \\
        &\_O\_\_ \\
        &P\_O\_ \\
        &and move 1 (Left) \\ 
        &resulted in no change in the board position, which indicates that the move was not successful due to the slippery ice.", "Win probability": "The win/success probability for P is not affected.", "Threat": "There are three holes on the board. There are two holes close to P, which can be dangerous if P moves right or up.", "Potential strategies": "Potential strategies for P include moving 4 (Up) to avoid the holes and move towards the goal position G."\} \\
        &"final\_evaluation": -0.4\} \\
        & \\
        &\#\#\# Evaluation for taking action 3: \\
        &\{"thought": \{"Reflection": "It appears that the initial board position \\
        &\_\_G\_ \\
        &\_\_\_O \\
        &\_O\_\_ \\
        &P\_O\_ \\
        &and move 3 (Right)  \\
        &was not favorable for P, as P is still far from the goal position G and there are holes nearby.", "Win probability": "The win/success probability for P is lower.", "Threat": "There are three holes on the board. There is a hole only one block away from P after action has been taken, which is very dangerous.", "Potential strategies": "Potential strategies for P include moving 4 (Up) to move closer to the goal position G." \} \\
        &"final\_evaluation": -0.5\} \\
        & \\
        & Now, please give your evaluation and the best next move based on the given board position \\
        & \_\_G\_ \\
        & \_\_\_O \\
        & \_O\_\_ \\
        & P\_O\_. \\
        &You should output your answer in the json format. Your answer consists of two elements: \\
        &- "thought": let's think step by step. Generate your detailed reflection by analyzing the next board positions and their evaluations. \\
        &- "best\_move": the best move for the next player. The move should be in the format of a number from 1 to 4, indicating the moving direction. \\
        &Don't output extra information except for the json format. \\
 \hline
ASSISTANT & \{ \\
  &"thought" : "Analyzing the given board position \_\_G\_ \_\_\_O \_O\_\_ P\_O\_, we can see that the goal is located at the top left corner, and there are three holes on the board. The player is currently at the bottom right corner. We have three possible moves: 4 (Up), 1 (Left), and 3 (Right). After evaluating the next board positions, we can see that moving 4 (Up) and 3 (Right) both result in a final evaluation of -0.5, indicating that the player is still far from the goal and there are holes nearby. Moving 1 (Left) results in no change in the board position due to the slippery ice, but the final evaluation is -0.4, which is slightly better than the other two options. However, considering the potential strategies, moving 4 (Up) seems to be a better option as it allows the player to move closer to the goal position G.", \\
  &"best\_move" : 4 \\
&\}\\
\hline
\end{longtable}
\end{mdframed}

\clearpage

\section{Prompts}\label{apx:prompt_details}
Here we provide all the prompt templates in three experiments.
\subsection{Maze experiment}
\begin{mdframed}[style=mystyle, frametitle={Language Value/Evaluation Prompt}]
\small
\begin{spverbatim}
EVAL_USER_PROMPT_S_V = f"""\
You are playing a simple maze game. An agent is moving in the maze and the objective is to reach the goal in as few steps as possible. 

The possible actions are "move up\n", "move down\n", "move left\n", "move right\n".

You are a highly skilled evaluator in this game and is expected to function similar to state value function in reinforcement learning. At each step I will provide you with the move history of the agent (from old to new), including where the goal is, your current position, the walls that surround the agent. Your output evaluation should be a json array that includes the following concepts:
- "thoughts": Let's think step by step. Generate your detailed thought process and evaluation of the agent's position and the distance towards the goal.
- "final_evaluation": Concisely judge how good the agent's current position is compared to all the possible positions in the maze, in terms of reaching the goal.

Let's start a new game. Now, please give your evaluation of the current state given the move history of the agent:
```
{{game_content}}
```

"""
\end{spverbatim}
\end{mdframed}
\begin{mdframed}[style=mystyle, frametitle={Language TD Prompt}]
\small
\begin{spverbatim}
EVAL_USER_PROMPT_S_TD_G2_new = f"""\
You are playing a simple maze game. An agent is moving in the maze and the objective is to reach the goal in as few steps as possible. 

The possible actions are "move up\n", "move down\n", "move left\n", "move right\n".

You are a highly skilled evaluator in this game, particularly adept at making accurate assessments through look-ahead of the current maze position after taking the given action. At each step I will provide you with the move history of the agent (from old to new), including where the goal is, your current position, the walls that surround the agent, the action that the agent is going to take, *along with several key variations of trajectory pieces after taking this action (and the corresponding natural language evaluations of the trajectory pieces)*. 
Your task is to understand these look-ahead information and summarize, derive non-trivial analysis and understanding the *the agent's position after taking the action*. Your output evaluation should be a json array with the following *two* concepts:
- "thoughts": Let's think step by step. Summarize the look-ahead information of the variations after taking action {{chosen_action}}.
- "final_evaluation": Now Concisely judge how good the chosen action is, in terms of reaching the goal.

Now, please give your evaluation given action {{chosen_action}}, the *current environment state*:
```
{{game_content}}
```

and the look-ahead information of different variations after taking action {{chosen_action}}:
"""
\end{spverbatim}
\end{mdframed}
\begin{mdframed}[style=mystyle, frametitle={Policy Improvement Prompt}]
\small
\begin{spverbatim}
POLICY_IMPROVEMENT_PROMPT_TD = f"""\
You are playing a simple maze game. An agent is moving in the maze and the objective is to reach the goal in as few steps as possible. 

Your task is to determine the best action for the next time step given the current state (the move history of the agent (from old to new), including where the goal is, your current position, the walls that surround the agent).

Your possible actions are "move up\n", "move down\n", "move left\n", "move right\n".

The evaluations of the agent after possible actions are given. Each of them consists of two elements:
- "thoughts": Summarization of the look-ahead information of the variations after taking the chosen action.
- "final_evaluation": Judge how good the chosen action is, in terms of reaching the goal.

DO NOT judge the action based on your exterior knowledge, only use the given evaluations to determine the best move.

Here are the evaluations of each possible action:

For action "move up", {{evaluations_up}}
,
For action "move down", {{evaluations_down}}
,
For action "move left", {{evaluations_left}}
,
For action "move right", {{evaluations_right}}

Return the best action (choose only one from the possible actions) given the evaluations in a json array with a key "action".
"""
\end{spverbatim}
\end{mdframed}
\subsection{Breakthrough Experiment}
\begin{mdframed}[style=mystyle, frametitle={Language Value/Evaluation Prompt}]
\addcontentsline{toc}{paragraph}{Language Value/Evaluation Prompt}
\small\ttfamily
\begin{spverbatim}
GAME_RULE_PROMPT = """Here is the rule for the Breakthrough board game:
The game is played on an 5x5 board for 2 players (white and black), with each player starting with 10 pawns. white pawns are on the first two rows and black pawns are on the last two rows.
Black moves first. In each turn, players can move one of their pieces one square forward, diagonally forward if the target square is empty. Or it can capture an opponent's piece if that square is one step diagonally forward.
The game ends when one player successfully break through the opponent lines -- either move a piece to the opposite last row of the board or captures all of the opponent's pieces.

For board representation, we use b for black pieces, w for white pieces, and . for empty squares. (1-5) and (a-e) are used to show the rows and columns index respectively."""

EVAL_SYSTEM_PROMPT = f"""{GAME_RULE_PROMPT}

You are a highly skilled evaluator in this game. I will provide you with specific board information representing the current board. Your output should include the following concepts:
1. *Tactical Considerations*: Any immediate threats, potential tactics, or vulnerabilities in the position.
2. *Positional Evaluation*: Consideration of pawn structure, piece activity, control of key squares, and game safety.
3. *Suggested Moves*: One or two strong candidate moves for the side to move, along with a brief rationale for comparing different moves.
4. *Advantage*: Based on all previous rationale, determine if white or black takes advantage. Use <white> or <black> to represent the winning side.

Your response should be informative and concise."""

EVAL_USER_PROMPT = """*The board you need to evaluate:* 
{board}"""
\end{spverbatim}
\end{mdframed}
\begin{mdframed}[style=mystyle, frametitle={Language TD Prompt}]
\addcontentsline{toc}{paragraph}{Language TD Prompt}
\small\ttfamily
\begin{spverbatim}
TD_SYSTEM_PROMPT = f"""{GAME_RULE_PROMPT}

You are a highly skilled evaluator in this game, particularly adept at making accurate assessments through look-ahead analysis of the current board position.
I will provide you with current board representation, *along with several key variations starting from this position (and their corresponding natural language evaluations of the subsequent positions)*.

Your task is to aggregate and compare these look-ahead information, to summarize, derive non-trivial analysis about the *current board*. Your output should include the following concepts:
1. *Tactical Considerations*: Any immediate threats, potential tactics, or vulnerabilities in the position.
2. *Positional Evaluation*: Consideration of pawn structure, piece activity, control of key squares, and game safety.
3. *Suggested Moves*: One or two strong candidate moves for the side to move, along with a brief rationale for comparing different moves.
4. *Advantage*: Based on all previous rationale, determine if white or black takes advantage. Use <white> or <black> to represent the winning side.

Your response should be informative and concise."""

TD_USER_PROMPT = """*The board you need to evaluate:* 

{board}

Here are the look-ahead variations from the current board position:
*Key Variations and Subsequent Evaluation:*:

{variations}

Please provide your analysis and understanding of the current board position based on the provided look-ahead information.
Your response should be informative and concise."""

VARIATION_PROMPT = """*Variation {i}:* 
Description of variation's move sequence:
{move_desc}

Subsequent position evaluation:
{subsequent_eval}"""

SUBSEQUENT_PROMPT = """The subsequent board is: 

{sub_board}

The evaluation of this subsequent board is: 

{sub_eval}"""
\end{spverbatim}
\end{mdframed}
\subsection{Tic-tac-toe Experiment}\label{app:prompts:tictactoe}

\begin{mdframed}[style=mystyle, frametitle={Policy Inference Prompt}]
\addcontentsline{toc}{paragraph}{Policy Inference Prompt}
\small\ttfamily
\begin{spverbatim}
POLICY_INFERENCE_SYSTEM_PROMPT = """\
You are an expert agent at playing the game tic-tac-toe on a 3*3 board. Tic Tac Toe is a two-player game played on a grid.
Players take turns marking a space with their respective symbol. The goal is to get multiple of one’s own symbols in a row, either horizontally, vertically, or diagonally, before the opponent does. If all nine squares are filled and no player has three in a row, the game is a draw.

Your task is to choose the best move given board position.
The board consists of "O", "X" and grid number. The grid number indicates empty grid.
You should output your answer in the json format. Your answer consists of two elements:
- "thought": let's think step by step. Generate your detailed thought process and plan for the next move.
- "best_move": the best move for the next player. The move should be in the format of a number from 1 to 9, indicating the position on the board. And the move should be in the available positions."""

POLICY_INFERENCE_EXAMPLE_USER_PROMPT = """\
Here is the board position and the next player is O:
O | O | X
---------
4 | X | 6
---------
7 | 8 | 9
The available move positions are 4, 6, 7, 8, 9.
"""

POLICY_INFERENCE_EXAMPLE_ASSISTANT_PROMPT = """\
{"thought": "It appears that the initial board position
O | O | X
---------
4 | X | 6
---------
7 | 8 | 9 was favorable for X, as X has occupied the positions 3, 5 and X can win by occupying the position 7. O has occupied the positions 1, 2. Therefore, the best move for O is to occupy the position 7 to block X and create a potential winning opportunity by occupying the positions 1, 4, 7.", "best_move": 7}
"""

POLICY_INFERENCE_USER_PROMPT = """\
Here is the board position and the next player is {next_player}:
{state}. The available move positions are {available_positions}.
"""
\end{spverbatim}
\end{mdframed}

\begin{mdframed}[style=mystyle, frametitle={Policy Improvement Prompt}]
\addcontentsline{toc}{paragraph}{Policy Improvement Prompt}
\label{app:prompts:tictactoe:policy-evaluation-prompt}
\small\ttfamily
\begin{spverbatim}
POLICY_IMPROVEMENT_SYSTEM_PROMPT = """\
You are playing the game tic-tac-toe on a 3*3 board. Tic Tac Toe is a two-player game played on a grid.
Players take turns marking a space with their respective symbol. The goal is to get multiple of one’s own symbols in a row, either horizontally, vertically, or diagonally, before the opponent does. If all nine squares are filled and no player has three in a row, the game is a draw.
The board consists of "O", "X" and grid number. The grid number indicates empty grid.
Your task is to determine the best move for the next player based on the given board position and the next player.
The evaluations of (board, action) pairs after possible moves are given.
DO NOT judge the board based on your knowledge, only use the evaluations to determine the best move.
The evaluation for the next board is in the format of a json format, consisting of two elements:
- "thought": evaluation of the the board and action pair.
- "final_evaluation": Judge which side takes advantage. 1 means 'O' takes advantage, -1 means 'X' takes advantage, and 0 means the game will be a draw.
"""

POLICY_IMPROVEMENT_USER_PROMPT = """\
Here is the board position and the next player is {next_player}:
{state}. The possible moves are {available_positions}.
The following are the boards after each possible move:
{next_states}

Now, please give your evaluation and the best move for {next_player} based on the given board position {state}.
You should output your answer in the json format. Your answer consists of two elements:
- "thought": let's think step by step. Generate your detailed reflection by analyzing the next board positions and their evaluations.
- "best_move": the best move for the next player. The move should be in the format of a number from 1 to 9, indicating the position on the board. And the move should be in the available positions {available_positions}.
Don't output extra information except for the json format.
"""
\end{spverbatim}
\end{mdframed}

\begin{mdframed}[style=mystyle, frametitle={Policy Evaluation Prompt}]
\addcontentsline{toc}{paragraph}{Policy Evaluation Prompt}
\small\ttfamily
\begin{spverbatim}
POLICY_EVALUATION_SYSTEM_PROMPT = """You are a player of the game of Tic Tac Toe. \nThe game goal is to get multiple of one's own symbols in a row, either horizontally, vertically, or diagonally, before the opponent does. If all nine squares are filled and no player has three in a row, the game is a draw. \nThe board consists of \"O\", \"X\" and grid number. The grid number indicates empty grid. \nYou are learning how to evaluate a (board, action) pair in the tic tac toe by playing the game given the (board, action) pair and reflect the playing history. \nThe playing history depicts a game in progress with a final result. Your answer consists of two elements:
- "thought": let's think step by step. Generate your detailed evaluation over the (board, action) pair by merely reflecting the playing history after this pair from different perspectives. You should only rely on the playing history as context and don't evaluate game with your own judgement. Your evaluation should contain the following elements: Win probability, Threat, Potential strategies.
- "final_evaluation": After all of your thoughts, judge which side takes advantage. 1 means you think 'O' takes advantage, -1 means you think 'X' takes advantage, and 0 means you think the game will be a draw.
You should output your answer in the json format."""
POLICY_EVALUATION_EXAMPLE_USER_PROMPT = """The (board, action) to evaluate is O's turn:
Board:
O | O | X
---------
4 | X | 6
---------
7 | 8 | 9
Action:
The O's move is 4.

Below is the rollout sequence after this (board, action):
After O's move 4, the board position is:
O | O | X
---------
O | X | 6
---------
7 | 8 | 9
After X's move 7, the board position is:
O | O | X
---------
O | X | 6
---------
X | 8 | 9
The game is over. X wins. X wins by occupying the positions 3, 5, 7."""

POLICY_EVALUATION_EXAMPLE_ASSISTANT_PROMPT = """
{"thought": {"Reflection": "It appears that the initial board position
O | O | X
---------
4 | X | 6
---------
7 | 8 | 9
and O's move 4 were not favorable for O, as X was able to block on O's move at 7 and ultimately win the game.", "Win probability": "The win probability for X is large, while the win probability for O is low.", "Threat": "X has played at 5 and 3. X can win by move 7. O can occupy 1, 4, 7, and create a potential winning opportunity. X occupies 5, which is a key position to win the game.", "Potential strategies": "Potential strategies for O include playing at 7 to block X and create a potential win by occupying 1, 4, 7. X could have occupied 3, 5, 7 to win the game. X has already occupied 3, 5, and needs only 1 move to complete the win."}
"final_evaluation": -0.8}"""
\end{spverbatim}
\end{mdframed}

\begin{mdframed}[style=mystyle, frametitle={Policy Evaluation Prompt (Continue)}]
\addcontentsline{toc}{paragraph}{Policy Evaluation Prompt}
\small\ttfamily
\begin{spverbatim}
POLICY_EVALUATION_USER_PROMPT = """The board to evaluate is {player}'s turn:
Board:
{board}
Action: The {player}'s move is {action}.

Below is the rollout sequence 1 after this (board, action):
After {player} taking action {action}, the board position is:
{rollout_board_1}
The game is over. {winner_1} wins. {winner_1} wins by occupying the positions {positions_1}.

...

Below is the rollout sequence N after this (board, action):
After {player} taking action {action}, the board position is:
{rollout_board_N}
The game is over. {winner_N} wins. {winner_N} wins by occupying the positions {positions_N}.

Now generate your evaluation for the (board, action) pair."""
\end{spverbatim}
\end{mdframed}

\section{Analogy}
\label{apx:analogy}
\subsection{Language MC}
\begin{figure}[t]
    \centering
    \includegraphics[width=\linewidth]{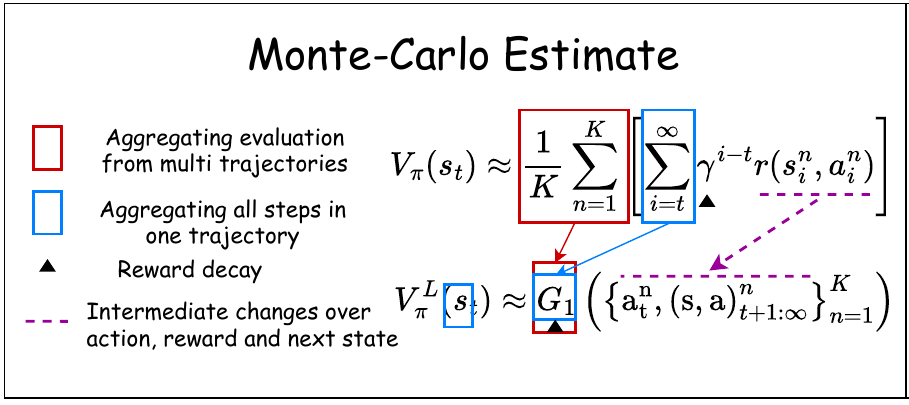}
    \caption{Comparison between traditional MC and language MC}
    \label{fig:mc}
\end{figure}
As mentioned in the main paper and in \cref{fig:mc}, language MC is an analogy of traditional MC. Specifically, the mean and sum operation in traditional MC estimate can be replaced by the new lew language aggregator $G_{1}$. $r$ can correspond to the intermediate transition for action, reward, and next state. The discount factor $\gamma$ can also have an interesting analogy in $G_{1}$ if $G_{1}$ acts as a lossy information compressor. Specifically, just as the discount factor reduces the weight of future rewards, $G_{1}$ can perform a lossy compression of future information, selectively retaining only the recent and most relevant aspects.
\subsection{Language TD}
\begin{figure}[t]
    \centering
    \includegraphics[width=\linewidth]{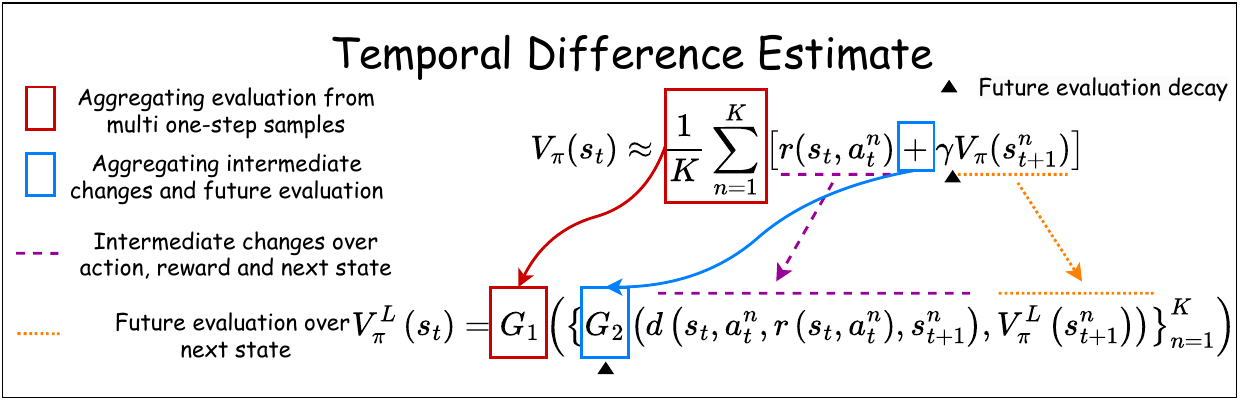}
    \caption{Comparison between traditional TD and language TD}
    \label{fig:td}
\end{figure}
Most Analogies between language TD and traditional TD share similarities with that in the language MC setting, as shown in \cref{fig:td}. $G_{1}$ is a language aggregator over multiple lookahead variations, corresponding to the trajectory mean operation, while $G_{2}$ combines immediate change/intermediate transition and future value evaluation, akin to the sum operation.

\section{Limitation and future work}\label{apx:limitation_and_future_work}

\paragraph{Limitations.} Currently, NLRL still faces several limitations. First, our experiments demonstrate its effectiveness primarily in environments with discrete action spaces and low-dimensional (textualisable) states. Its performance in environments involving continuous action spaces and high-dimensional states (e.g., low-level robotic control tasks) remains largely unexplored. Second, our experiments are primarily conducted in small-scale settings due to the substantial computational cost associated with invoking Large Language Models. The time efficiency of the current NLRL approach is significantly lower than that of traditional RL methods using smaller networks.

\paragraph{Future work.} Given NLRL's potential, there are several promising directions for future research. First, a theoretical foundation for the entire framework in the language space is needed to move beyond analogical reasoning towards a more robust, principled basis. This would help formalise the framework and enhance its generalisability. Another important direction is exploring how NLRL can be integrated more organically with existing research on self-evaluation, self-improvement, LLM agent framework, and LLM's planning. Such integration could unlock new capabilities and efficiency improvements. Additionally, in terms of application domains, there is significant potential to extend NLRL beyond the current setups--we are actively exploring extending NLRL to general domains beyond games, such as reasoning tasks \citep{cobbe2021gsm8k} and code \citep{chen2021evaluating}.

%%%%%%%%%%%%%%%%%%%%%%%%%%%%%%%%%%%%%%%%%%%%%%%%%%%%%%%%%%%%

\newpage
\section*{NeurIPS Paper Checklist}

\begin{enumerate}

\item {\bf Claims}
    \item[] Question: Do the main claims made in the abstract and introduction accurately reflect the paper's contributions and scope?
    \item[] Answer: \answerYes{} % Replace by \answerYes{}, \answerNo{}, or \answerNA{}.
    \item[] Justification: Section \ref{sec:exp} and Appendix \ref{apx:exp} have discussed related experiments claimed in the abstract and introduction in full detail. The conclusion supports the paper's contributions and scope.

\item {\bf Limitations}
    \item[] Question: Does the paper discuss the limitations of the work performed by the authors?
    \item[] Answer: \answerYes{} % Replace by \answerYes{}, \answerNo{}, or \answerNA{}.
    \item[] Justification: Appendix \ref{apx:limitation_and_future_work} has discussed potential limitation of our works.

\item {\bf Theory assumptions and proofs}
    \item[] Question: For each theoretical result, does the paper provide the full set of assumptions and a complete (and correct) proof?
    \item[] Answer: \answerNA{} % Replace by \answerYes{}, \answerNo{}, or \answerNA{}.
    \item[] Justification: The paper does not include theoretical results.

    \item {\bf Experimental result reproducibility}
    \item[] Question: Does the paper fully disclose all the information needed to reproduce the main experimental results of the paper to the extent that it affects the main claims and/or conclusions of the paper (regardless of whether the code and data are provided or not)?
    \item[] Answer: \answerYes{} % Replace by \answerYes{}, \answerNo{}, or \answerNA{}.
    \item[] Justification: Section \ref{sec:exp}, Appendix \ref{apx:exp} and \ref{apx:prompt_details} include detailed experiments setup and hyperparameter settings used in our work.

\item {\bf Open access to data and code}
    \item[] Question: Does the paper provide open access to the data and code, with sufficient instructions to faithfully reproduce the main experimental results, as described in supplemental material?
    \item[] Answer: \answerYes{} % Replace by \answerYes{}, \answerNo{}, or \answerNA{}.
    \item[] Justification: The paper provides open access to the code anonymously.

\item {\bf Experimental setting/details}
    \item[] Question: Does the paper specify all the training and test details (e.g., data splits, hyperparameters, how they were chosen, type of optimizer, etc.) necessary to understand the results?
    \item[] Answer: \answerYes{} % Replace by \answerYes{}, \answerNo{}, or \answerNA{}.
    \item[] Justification: Section \ref{sec:exp}, Appendix \ref{apx:exp} and \ref{apx:prompt_details} include detailed experiments setup and hyperparameter settings used in our work.

\item {\bf Experiment statistical significance}
    \item[] Question: Does the paper report error bars suitably and correctly defined or other appropriate information about the statistical significance of the experiments?
    \item[] Answer: \answerYes{} % Replace by \answerYes{}, \answerNo{}, or \answerNA{}.
    \item[] Justification: The paper reports error bars on some of the experiments, such as Table \ref{tab:language_td}. Due to high computational burdens, the paper does not include error bars for all of our experiments, but the paper contains every detail necessary for experiment reproduction.

\item {\bf Experiments compute resources}
    \item[] Question: For each experiment, does the paper provide sufficient information on the computer resources (type of compute workers, memory, time of execution) needed to reproduce the experiments?
    \item[] Answer: \answerYes{} % Replace by \answerYes{}, \answerNo{}, or \answerNA{}.
    \item[] Justification: The paper includes the computational resources required to run the experiments in Appendix \ref{apx:exp}.
    
\item {\bf Code of ethics}
    \item[] Question: Does the research conducted in the paper conform, in every respect, with the NeurIPS Code of Ethics \url{https://neurips.cc/public/EthicsGuidelines}?
    \item[] Answer: \answerYes{} % Replace by \answerYes{}, \answerNo{}, or \answerNA{}.
    \item[] Justification: The paper conforms with the Code of Ethics.

\item {\bf Broader impacts}
    \item[] Question: Does the paper discuss both potential positive societal impacts and negative societal impacts of the work performed?
    \item[] Answer: \answerNA{} % Replace by \answerYes{}, \answerNo{}, or \answerNA{}.
    \item[] Justification: The paper mainly focuses on foundational research and and not tied to particular applications.
    
\item {\bf Safeguards}
    \item[] Question: Does the paper describe safeguards that have been put in place for responsible release of data or models that have a high risk for misuse (e.g., pretrained language models, image generators, or scraped datasets)?
    \item[] Answer: \answerNA{} % Replace by \answerYes{}, \answerNo{}, or \answerNA{}.
    \item[] Justification: The paper does not release models or datasets.

\item {\bf Licenses for existing assets}
    \item[] Question: Are the creators or original owners of assets (e.g., code, data, models), used in the paper, properly credited and are the license and terms of use explicitly mentioned and properly respected?
    \item[] Answer: \answerYes{} % Replace by \answerYes{}, \answerNo{}, or \answerNA{}.
    \item[] Justification: Existing assets used in the paper are credited and the necessary licenses are mentioned. 

\item {\bf New assets}
    \item[] Question: Are new assets introduced in the paper well documented and is the documentation provided alongside the assets?
    \item[] Answer: \answerYes{} % Replace by \answerYes{}, \answerNo{}, or \answerNA{}.
    \item[] Justification: The new assets introduced is well documented.

\item {\bf Crowdsourcing and research with human subjects}
    \item[] Question: For crowdsourcing experiments and research with human subjects, does the paper include the full text of instructions given to participants and screenshots, if applicable, as well as details about compensation (if any)? 
    \item[] Answer: \answerNA{} % Replace by \answerYes{}, \answerNo{}, or \answerNA{}.
    \item[] Justification: The paper does not involve crowdsourcing nor research with human subjects.

\item {\bf Institutional review board (IRB) approvals or equivalent for research with human subjects}
    \item[] Question: Does the paper describe potential risks incurred by study participants, whether such risks were disclosed to the subjects, and whether Institutional Review Board (IRB) approvals (or an equivalent approval/review based on the requirements of your country or institution) were obtained?
    \item[] Answer: \answerNA{} % Replace by \answerYes{}, \answerNo{}, or \answerNA{}.
    \item[] Justification: The paper does not involve crowdsourcing nor research with human subjects.

\item {\bf Declaration of LLM usage}
    \item[] Question: Does the paper describe the usage of LLMs if it is an important, original, or non-standard component of the core methods in this research? Note that if the LLM is used only for writing, editing, or formatting purposes and does not impact the core methodology, scientific rigorousness, or originality of the research, declaration is not required.
    %this research? 
    \item[] Answer: \answerYes{} % Replace by \answerYes{}, \answerNo{}, or \answerNA{}.
    \item[] Justification: The paper is about foundational research on LLM and has described the usage of LLMs.

\end{enumerate}

\end{document}